%% file: root.tex
\documentclass[conference,a4paper]{IEEEtran}
\IEEEoverridecommandlockouts
\usepackage[pdftex]{graphicx}
\usepackage{amsmath,amssymb}
\usepackage{tikz}
\usepackage{pgfplots}
\usepackage{xspace}
\usepackage{tabularx}
\usepackage{textcomp}
\usepackage{gensymb}

\usepackage{multirow}
\pgfplotsset{compat=1.13}
\usepackage{url}
\usepackage{algorithmicx}
\usepackage{algorithm} 
\usepackage{algpseudocode} 

\algnewcommand\algorithmicinput{\textbf{Input: }}
\algnewcommand\Input{\item[\algorithmicinput]}
\algnewcommand\algorithmicoutput{\textbf{Output: }}
\algnewcommand\Output{\item[\algorithmicoutput]}
\algtext*{EndWhile}
\algtext*{EndFor}
\algtext*{EndIf}

\makeatletter
\DeclareRobustCommand\onedot{\futurelet\@let@token\@onedot}
\def\@onedot{\ifx\@let@token.\else.\null\fi\xspace}
\def\eg{\emph{e.g}\onedot} 
\def\ie{\emph{i.e}\onedot}

\def\etal{\emph{et al}\onedot}
\makeatother

\begin{document}
\title{Facetwise Mesh Refinement for Multi-View Stereo}

\author{
\IEEEauthorblockN{Andrea Romanoni}
\IEEEauthorblockA{Politecnico di Milano, Italy* \thanks{*Work done prior to Amazon involvement of the author and does not reflect views of the Amazon company}\\
andrea.romanoni@polimi.it}
\and
\IEEEauthorblockN{Matteo Matteucci}
\IEEEauthorblockA{Politecnico di Milano, Italy\\
matteo.matteucci@polimi.it}}

\maketitle
\begin{abstract}
Mesh refinement is a fundamental step for accurate Multi-View Stereo. 
It modifies the geometry of an initial manifold mesh to minimize the photometric error induced in a set of camera pairs.
This initial mesh is usually the output of volumetric 3D reconstruction based on min-cut over Delaunay Triangulations. Such methods produce a significant amount of non-manifold vertices, therefore they require a vertex split step to explicitly repair them.
In this paper, we extend this method to preemptively fix the non-manifold vertices by reasoning directly on the Delaunay Triangulation and avoid most vertex splits.
The main contribution of this paper addresses the problem of choosing the camera pairs adopted by the refinement process.
We treat the problem as a mesh labeling process, where each label corresponds to a camera pair.
Differently from the state-of-the-art methods, which use each camera pair to refine all the visible parts of the mesh,  we choose, for each facet, the best pair that enforces both the overall visibility and coverage. The refinement step is applied for each facet using only the camera pair selected. This facetwise refinement helps the process to be applied in the most evenly way possible. 
\end{abstract}

\IEEEpeerreviewmaketitle

\section{Introduction}
\label{sec:intro}
Recovering an accurate mesh representation of the world from a set of unordered images has been a long-standing issue in the Computer Vision community and goes under the name of Multi-View Stereo (MVS).
City mapping, cultural heritage preservation, object digitalization are some of its well-known applications.

Fig. \ref{fig:pipeline} illustrates the pipeline of a typical Multi-View Stereo system.
After camera poses are extracted via Structure from Motion \cite{wu2011visualsfm,schonberger2016structure,moulon2012adaptive}, MVS methods estimate a depth map for each image and fuse them into a 3D dense point cloud \cite{schonberger2016pixelwise,yao2018mvsnet,shen2013accurate}. From the reconstructed 3D points and the camera-to-point visibility, the successive volumetric 3D reconstruction step generates a 3D model in the form of a triangular mesh. To obtain high accurate and scalable results, volumetric methods based on graph-cut over the Delaunay Triangulation \cite{labatut2007efficient,tola2012efficient} are usually preferred over their voxel-based counterparts \cite{savinov2015discrete,hane_et_al_09,zach2007globally}.
The mesh created likely contains singular, \ie non-manifold, vertices that are undesirable for further mesh processing and for the successive mesh refinement step.

In this paper, we propose to preemptively avoid generating most of the singular vertices, by adjusting the free-space/matter labeling resulting from the graph-cut algorithm on the Delaunay Triangulation. This diminishes significantly the need for duplicating the non-manifold vertices, which is the usual procedure to hallucinate a manifold mesh.

The principal contribution of this paper focuses on mesh refinement, which is the last step of the MVS pipeline adopted to add details to the 3D model recovered by volumetric methods. 
Mesh refinement minimizes the photometric error induced by the 3D model among pairs of cameras.
State-of-the-art approaches choose these pairs relying on just the visibility of Structure from Motion 3D points \cite{vu_et_al_2012,li2016efficient}. Only in \cite{romanoni2019mesh} the richer visibility information carried out by the initial mesh is partially exploited.
In both cases, they use each pair to refine the whole visible part of the mesh giving the same relevance to all the cameras. Indeed, they pair each camera with at least one other, even if the camera carries redundant or poor information.
In this paper, we explicitly choose to refine each facet of the mesh with a specific pair of cameras. This addresses the problem of maximizing the surface coverage and its visibility and improving the accuracy of the refinement process. As a side effect, this method diminishes the rendering complexity.

\begin{figure*}[tbp]
    \centering
    \includegraphics[width=0.75\textwidth]{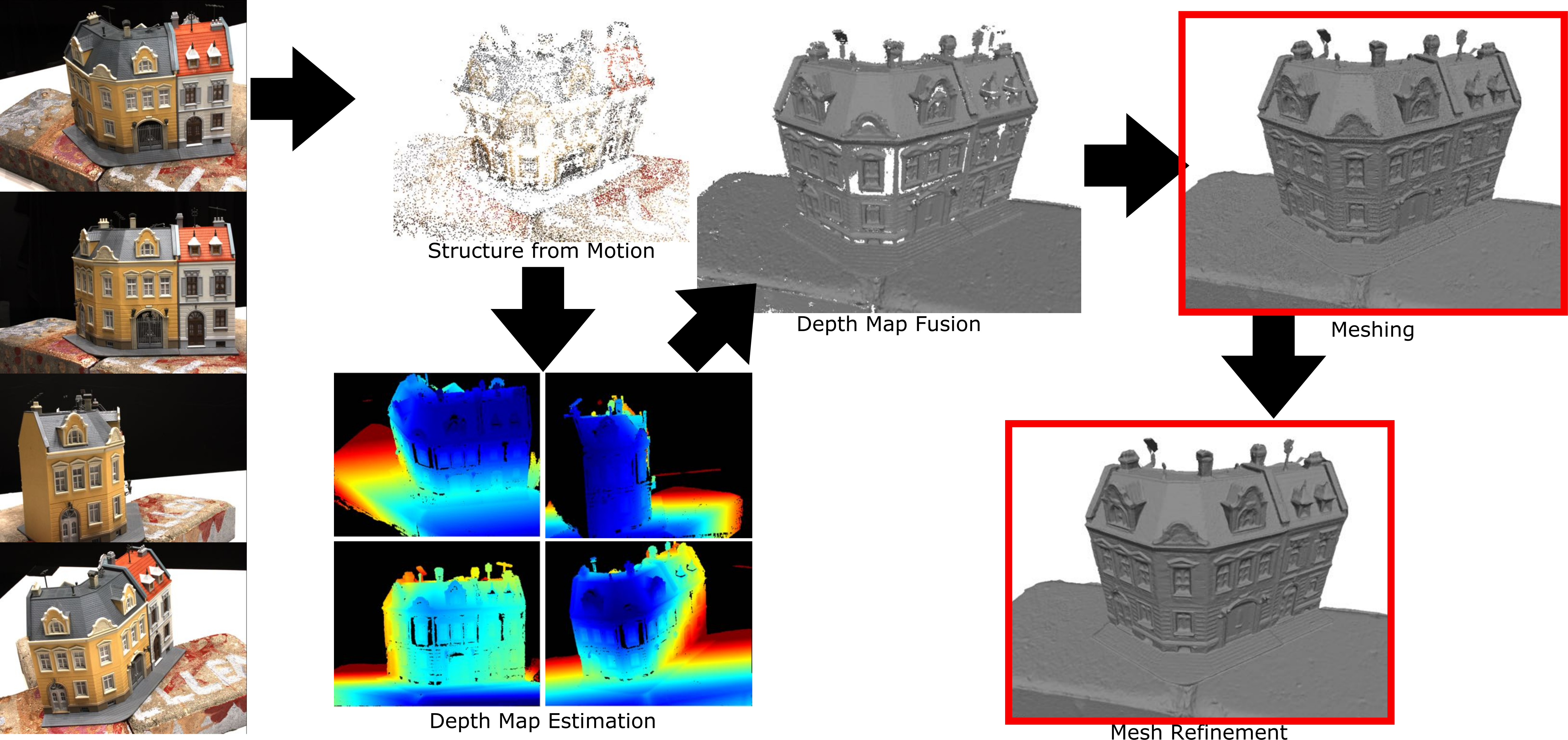}
    \caption{Typical pipeline of a Multi-View Stereo method. In this paper we focus on the meshing and the refinement steps}
    \label{fig:pipeline}
\end{figure*}

\section{Related works}
\label{sec:related}

In this section, we give a brief overview of the two steps covered in this paper, \ie volumetric 3D reconstruction and mesh refinement.

Volumetric 3D reconstruction is the most popular and effective approach to recover the 3D mesh of a scene.
The idea is to leverage ray visibility information to understand which part of the scene is \emph{free-space} and which part is \emph{matter}.
A successful class of volumetric methods \cite{savinov2015discrete,hane_et_al_09,zach2007globally} rely on voxel-based partitioning of the space. This approach leads to high-quality results but often suffers scalability issues. Moreover, to produce a surface mesh out of the volumetric representation, they apply the marching cubes algorithm \cite{lorensen1987marching} or Poisson Reconstruction \cite{kazhdan2006poisson} that generally outputs over-smoothed models, not capturing fine details of the scene.

On the other hand, several volumetric methods are based on Delaunay Triangulation \cite{labatut2007efficient,tola2012efficient,romanoni15b,romanoni16}.
Differently from the voxel-based, they are inherently more scalable, since Delaunay triangulation adapts its density according to the point cloud. Moreover, after the visibility is fused into the volume, no further meshing step is needed, indeed, the triangular mesh in output is simply the boundary between free and matter tetrahedra. 
Various approaches have been developed to define such free/matter partitioning. 
The simplest method initializes all the space as matter and marks as free-space each tetrahedron intersected by the camera-to-point rays\cite{lovi_et_al_11}. 
More advanced methods \cite{lhuillier_yu2013,litvinov_lhiuller14,romanoni15b} rely on region growing to explicitly enforces the 2-manifold property, while incrementally creating the mesh.
When an incremental reconstruction and the  2-manifold property are not needed but robust and accurate meshes from a very dense point cloud are required, graph-cut-based methods represent a more effective alternative \cite{labatut2007efficient,vu_et_al_2012,tola2012efficient}. In this case, the nodes of the graph correspond to tetrahedra and the edges to facets, \ie they encode the adjacency of the triangulation. By defining a convenient energy function over this graph, it is possible to extract the surface as the set of facets that corresponds to the minimum cut. 

To improve the reconstructed model, the final step of the MVS pipeline is mesh refinement. 
Early mesh refinement minimizes a photometric energy function defined over implicit surface representation \cite{jin2002variational,pons2007multi,yoon2010joint}.
More recent and effective methods directly evolve the surface to minimize the photometric error induced by the reprojection of an image to a reference image through the 3D model.
One of the most successful approaches was proposed by Vu \etal \cite{vu_et_al_2012}. This method has also been extended to exploit semantics \cite{blaha2017semantically,romanoni2017multi} or optimized with adaptive resolution \cite{li2016efficient}.

\begin{figure*}[t]
\centering
\setlength{\tabcolsep}{1px}
\begin{tabular}{cccc}
{\def\svgwidth{0.23\textwidth}
  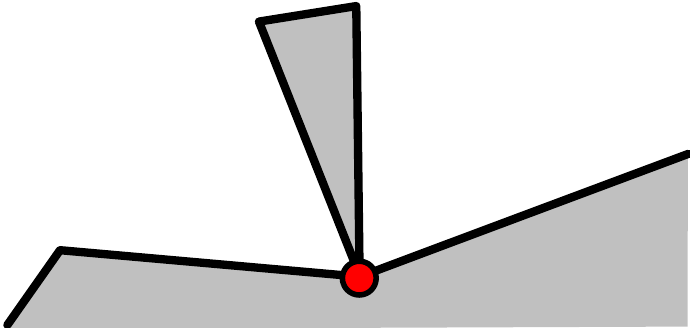}&
{\def\svgwidth{0.23\textwidth}
  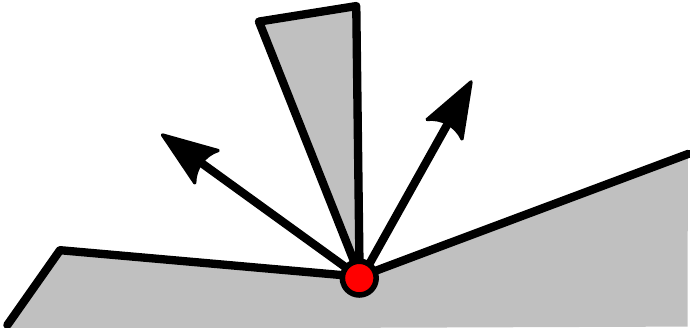}&
{\def\svgwidth{0.23\textwidth}
  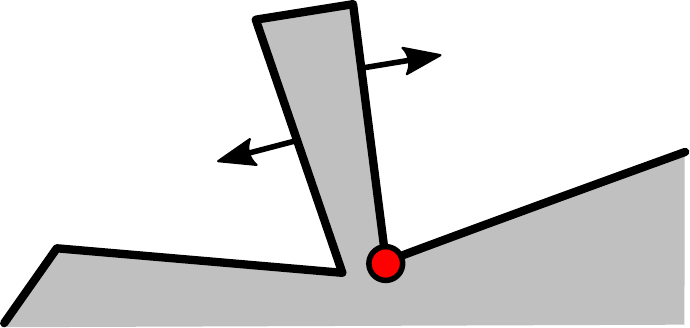}&
{\def\svgwidth{0.23\textwidth}
  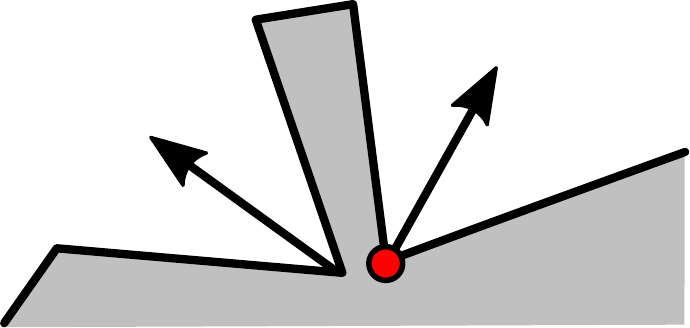}\\
before refinement&after refinement&before refinement&after refinement\\
\multicolumn{2}{c}{singular vertex}&
\multicolumn{2}{c}{manifold vertex}\\
\end{tabular}
\caption{Surface evolution of non manifold (a-b) and manifold (c-d) vertices. The arrows before refinement represent the normals pointing outwards the triangles, and after refinement represent the surface evolution flow.}
\label{fig:whymanifold}
\end{figure*}

\subsection{Mesh Refinement}
\label{subsec:refinement}
In this Section, we give a brief overview of the mesh refinement problem formulated in \cite{pons2007multi}  then extend in \cite{vu_et_al_2012} for more in-depth analysis and more explanatory figures we refer the reader to these papers. Note that this summary partially overlaps with those presented in a previous work \cite{romanoni2019mesh} describing the same background concepts. 

Given a set of images and an initial mesh, the idea is to minimize the energy 
$
E = E_{\textrm{photo}} + E_{\textrm{smooth}}.
$

The energy $E_{\textrm{photo}}$ is related to the photometric error induced by the model:
\begin{equation}
\label{eq:energy_photo}
  E_{\textrm{photo}} = E(\mathit{S})  = \sum_{i,j}\int_{\Omega^{\textrm{S}}_{i,j}} err_{I, I_{ij}^{\mathit{S}}}(x)\textrm{d}x   = \sum_{i,j} \mathcal{E}^{im}_{ij}(x),
\end{equation}
where $\mathit{S}$ is a generic 2D surface embedded in 3D, $err_{I, I_{ij}^{\mathit{S}}}(x)$ represents the photometric error between the patch around the projection of the 3D point $x\in\mathit{S}$ in $I$ and $I_{ij}^{\mathit{S}}$. In our case, the error function is the negative ZNCC of the $5x5$ pixels neighborhood. $I_{ij}^{\mathit{S}}$ represents the reprojection of the image from the $j$-th camera onto camera $i$ ,\ie, image $I$, through the surface $\mathit{S}$, and $\Omega^{\textrm{S}}_{i,j}$ is the image region where the reprojection is defined.

Since the surface  $\mathit{S}$ is a mesh model,  we compute the discrete gradient of $E_{\textrm{photo}}$ with respect to the vertex $V \in \mathit{S}$:
\begin{align}
\begin{split}
  \frac{\textrm{d}E(\mathit{S})}{\textrm{d}V}& =  \int_{\mathit{S}} \phi_V(x) \nabla E(S) \textrm{d}x =
  \int_{\mathit{S}} \phi_V(x) \sum_{i,j} \nabla \mathcal{E}^{im}_{ij}(x)\textrm{d}x.
\end{split}
\end{align}
where  $\phi_V(x)$ is the barycentric coordinate of a 3D point $x$ belonging to the triangle with vertex $V$ of the surface $\mathit{S}$.
It is possible to integrate the energy more easily over the image by changing the variable of integration  with $\textrm{d}x_i = -\overrightarrow{n}^T \mathbf{d}_i \textrm{d}x/z_i^3$ (see  \cite{pons2007multi}), where $\overrightarrow{n}$ is the normal at $x$ pointing outward $\mathit{S}$, $x_i$ the projection of x into the  image $I$, $\mathbf{d}_i$ is the vector from camera $i$ to $x$, $z_i$ is the depth of $x$ in camera $i$. So we obtain
\begin{equation}
\label{eq:final}
  \frac{\textrm{d}E(\mathit{S})}{\textrm{d}V} = 
  \sum_{i,j} \int_{\Omega^{\textrm{S}}_{i,j}} 
  \phi_V(x)  \nabla \mathcal{E}^{im}_{ij}(x)\frac{z_i^3}{\overrightarrow{n}^T \mathbf{d}_i }\overrightarrow{n} \textrm{d}x_i.
\end{equation}
The mesh refinement algorithm therefore collects, for each vertex, the gradient contributions computed for each point of the adjacent triangles, weighted by their barycentric coordinates.

Then, to minimize the energy term $E_{\textrm{smooth}}$, the evolution is complemented by applying the umbrella operator \cite{wardetzky2007discrete}. It approximates the Laplace-Beltrami operator by moving each vertex in the mean position of its neighbors.

One of the open challenges in mesh refinement is how to choose conveniently the camera pairs $(i,j)$ that we are going to compare to compute the photometric error.
Often classical Multi-View Stereo methods \cite{tola2012efficient,vu_et_al_2012,VuPhD011} leverage  3D points correspondences estimated by the Structure from Motion. 
Therefore, for each $i$, the camera $j$ with the highest number of common 3D points with a reasonable parallax (\eg, in \cite{tola2012efficient} between $10\degree$ and $30\degree$) is selected.
Recently, Romanoni and Matteucci \cite{romanoni2019mesh} proposed to exploit the knowledge of the model to compute an energy term that considers image overlap, parallax, symmetry, and coverage to evaluate, for each camera $i$ the best $j$ to be compared to.

A relevant limitation of these approaches is that, once the subset of camera pairs is selected they give essentially the same relevance to all of them. 
In this paper instead, we change the perspective. We apply a facetwise refinement, \ie the contribution of each facet to Eq. \eqref{eq:final} is computed using only the camera pair providing the best visibility.

\begin{figure*}[tbp]
    \centering
    \setlength{\tabcolsep}{1px}
    \begin{tabular}{ccc}
        \def\svgwidth{0.25\textwidth}
        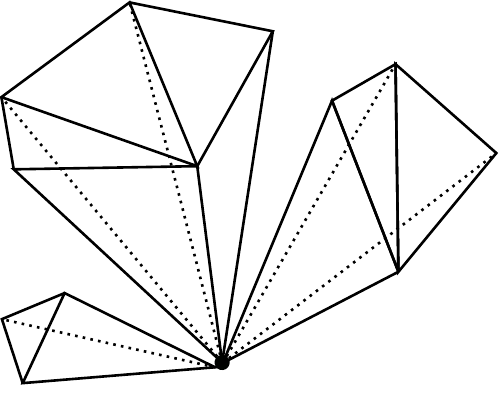
        & 
        \def\svgwidth{0.25\textwidth}
        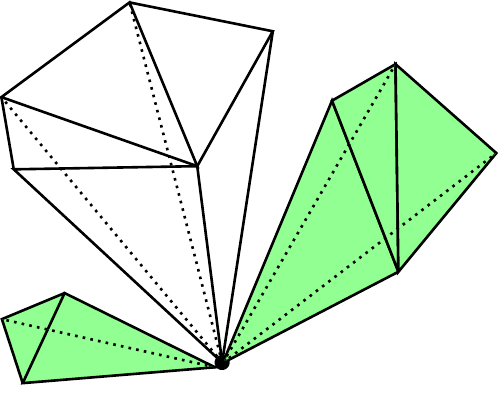
        & 
        \def\svgwidth{0.25\textwidth}
        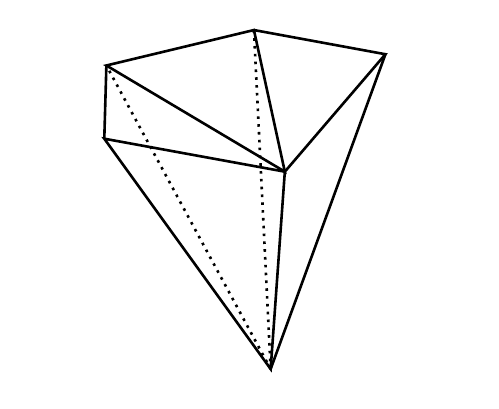\\
        (a)&(b)&(c)
    \end{tabular}
   
    \caption{Fixing the singular vertices in the Delaunay triangulation. The connected components depicted belong to matter, while the free-space are the remaining incident tetrahedra we do not depict to simplify the drawing. Green tetrahedra are those re-labeled from matter to free-space.}
    \label{fig:connected}
\end{figure*}

\subsection{2-manifold property}
\label{subsec:manifold}
The manifold property is necessary to apply coherently both the mesh refinement and other Computer Graphics algorithms, which improve the quality of the model.
A mesh is manifold if every vertex $v$, its neighboring vertices are homeomorphic to a disk, \ie they form a single unique cycle around the $v$. The vertices breaking the manifold property are named \emph{singular vertices}.

In Fig. \ref{fig:whymanifold}, we show the importance of the manifold property for mesh refinement. In the left, the vertex $v$ is singular, Equation \eqref{eq:final} for the vertex $v$ sums up the gradient contribution carried by the triangles $T_1$, $T_2$, $T_3$ and $T_4$ (Fig. \ref{fig:whymanifold}(b)); since the corresponding normals have very different directions, these contributions and the resulting vector flows contain contradictory information.
On the right, instead, the vertices are manifold and the gradients collected in $v_1$ and $v_2$ (Fig. \ref{fig:whymanifold}(d)) lead to a coherent mesh evolution.

\section{System Overview}
\label{sec:overview}
The proposed pipeline takes as input a set of RGB images and produces an accurate 3D model mesh of the scene. All the steps involved do not require any human intervention especially thanks to the meshing step which outputs a 2-manifold mesh.

We adopted COLMAP \cite{schonberger2016structure,schonberger2016pixelwise} to estimate the camera poses (Structure from Motion), to generate the depth-maps and normal-maps corresponding to each RGB image and to fuse those outputs into a single point cloud keeping the mutual coherence of the depth and normal estimates across the views. 
From the extracted point cloud we generate a manifold mesh with our extension of the graph-cut meshing step proposed by Labatut \etal \cite{labatut2007efficient} that ensures the surface to be 2-manifold.
Finally, we exploit facet visibility to define the set of cameras that we successively use for the last mesh refinement step.

In the following, we focus on the 2-manifold meshing and the facetwise mesh refinement. We refer the reader to \cite{schonberger2016pixelwise} for a better understanding of the depth-map estimation and fusion steps.

\section{2-manifold meshing}
\label{sec:manifold}
To reconstruct the 3D model that is refined in the next step of the pipeline, we extend the method proposed by Labatut \etal \cite{labatut2007efficient}.
From the set of 3D points extracted after depth map fusion and their respective visibility rays from the cameras, as in \cite{labatut2007efficient}, we first compute their 3D Delaunay Triangulation.
They create a graph with a node corresponding to each tetrahedron and an edge corresponding to each face connecting adjacent tetrahedra. 
For each camera-point visibility ray, the tetrahedron where the camera is located is connected to the source node with an edge of weight $\alpha_{\textrm{vis}}$.
The weight of the edges corresponding to the intersected facets are increased by $\alpha_{\textrm{vis}}(1-e^{d/2\sigma})$, where $d$ is the distance between the facet and the 3D point and $\sigma$ is the 25th percentile of all the edges inside the triangulation.
Finally, the tetrahedron behind the 3D point and with a distance of 3$\sigma$ is connected to the sink node. 
In addition to this visibility voting scheme, they also add a prior about the surface quality that favors smooth surfaces.

Successively the graph cut algorithm looks for the minimum cut and labels the tetrahedra either free-space and matter.
In Labatut \etal \cite{labatut2007efficient}  the boundary between the two sets is the final mesh.
However, such mesh usually contains a significant number of non-manifold vertices.

Lhuiller \cite{lhuillier2018surface} extends the definition of manifold surface to the case in which the triangular mesh is the boundary $\delta F$ between two disjoint sets of tetrahedra $\mathcal{F}$ and $\mathcal{M}$ (free-space and matter), as the output of Labatut's min-graph cut algorithm.
Let $T_v$ be the set of tetrahedra incident to a vertex $v \in \delta F$ of the mesh. 
$T_v$ can be split in connected components belonging to either $\mathcal{M}$ and $\mathcal{F}$. Fig. \ref{fig:connected} illustrates three components in  $\mathcal{M}$ and implies that the other tetrahedra are free, therefore in the example $\mathcal{F}$ is one single component. 
According to Theorem 3 in \cite{lhuillier2018surface}, $v$ is singular iff the number of connected components are three or more (as in Fig. \ref{fig:connected}(a)).

Therefore, for each vertex $v \in \delta F$ let us split the set of incident tetrahedra in two subsets of connected components: $\mathcal{CC}_v^{\mathcal{F}} =\{CC_1^{\mathcal{F}}, \cdots, CC_K^{\mathcal{F}}\}$ belong to ${\mathcal{F}}$ and $\mathcal{CC}_v^{\mathcal{M}} = \{CC_1^{\mathcal{M}}, \cdots, CC_L^{\mathcal{M}}\}$ belong to $\mathcal{M}$.
In case of $|\mathcal{CC}_v^{\mathcal{F}}|+|\mathcal{CC}_v^{\mathcal{M}}|>2$ the vertex is singular.

To preemptively avoid as many non-manifold vertices as possible, for each singular vertex we try to remove the source of non-manifoldness with the following iterative procedure (Fig. \ref{Algo3}). 
First, we collect all the singular vertices according to the definition above (Fig. \ref{Algo1}).
Then, for each singular vertices we
(i) compute the connected components of matter incident to $v$, set all tetrahedra to free space except to those belonging to the component with the highest cardinality (as the green one in Fig. \ref{fig:connected}(b) that turns into free-space in Fig. \ref{fig:connected}(c)). Then, do the same for the connected components of free space. 
(ii) update the list of singular vertices. (iii) compute the connected components of matter incident to $v$ and split at the centroid each tetrahedron except to those belonging to the component with the highest cardinality. Then, do the same for the connected components of free space. 
Finally, repeat update the singular vertices list and redo (i); the tetrahedron splits in (iii) help escaping local minima. The pseudocode of this procedure is in (Fig. \ref{Algo2}).

This process usually diminishes dramatically the singular vertices of the mesh extracted from the Delaunay Triangulation. However, sometimes it gets stuck in local minima. For this reason, we eventually split the few remaining singular vertices.



\begin{figure}
\begin{algorithmic}[1]
\Input{ $<T,V>$ 3D Delaunay Triangulation with tetrahedra in $T$ and vertices in $V$}
\Procedure{Cleanup Vertices Fixing}{}
    \State listSingularV = $\{\}$
    \For {$v$ in $V$}
        \If {IsSingular($T$,$v$)} listSingularV.push($v$)
        \EndIf
    \EndFor
    \State  Singular\_Vertex\_Fixing($listSingularV$,false,$T,V$)
    \State listSingularV = $\{\}$
    \For {$v$ in $V$}
        \If {IsSingular($T$,$v$)} listSingularV.push($v$)
        \EndIf
    \EndFor
    \State  Singular\_Vertex\_Fixing($listSingularV$,true,$T,V$)
    \State listSingularV = $\{\}$
    \For {$v$ in $V$}
        \If {IsSingular($T$,$v$)} listSingularV.push($v$)
        \EndIf
    \EndFor
    \State  Singular\_Vertex\_Fixing($T,V$,$listSingularV$,false)
\EndProcedure
\end{algorithmic}
\caption{Cleanup Singular vertices algorithm}
\label{Algo3}
\end{figure}

 \begin{figure}
 \begin{algorithmic}[1]
 \renewcommand{\algorithmicrequire}{\textbf{Input:}}
 \renewcommand{\algorithmicensure}{\textbf{Output:}}
 \Input{ $<T,V>$ 3D Delaunay Triangulation with tetrahedra in $T$ and vertices in $V$, vertex $v\in V$}
 \Output  {True if the vertex is singular, False otherwise}
\Procedure{IsSingular}{}
    \State $\mathcal{CC}_v^{\mathcal{F}}$ =  FreeConnectedComponents($T$,$v$)
    \State $\mathcal{CC}_v^{\mathcal{M}}$ =  MatterConnectedComponents($T$,$v$)
    \State\Return $|\mathcal{CC}_v^{\mathcal{F}}|+|\mathcal{CC}_v^{\mathcal{M}}|>2$
\EndProcedure
 \end{algorithmic}
 \caption{IsSingular algorithm.  FreeConnectedComponents and MatterConnectedComponents extract the list of set of connected tetrahedra}
 \label{Algo1}
 \end{figure}

\begin{figure}
\begin{algorithmic}[1]
\algrenewcommand\algorithmicindent{1.0em}%
\Input{ $listSingularV$ list of singular vertices, $SPLIT$ True to apply tetrahedra splitting, $<T,V>$ 3D Delaunay Triangulation}
\Procedure{Singular\_Vertex\_Fixing}{}
\For {$v$ in $listSingularV$}
    \If {IsSingular($v$)}
        \State $\mathcal{CC}_v^{\mathcal{M}}$ =  MatterConnectedComponents($T$,$v$)
        \State sort\_ascending\_cardinality($\mathcal{CC}_v^{\mathcal{M}}$)
        \State  //Set to free all but the CC with most tetrahedra
        \For {$i=0$ to $|\mathcal{CC}_v^{\mathcal{M}}|$-1}
            \For {$\Delta$ in $\mathcal{CC}_v^{\mathcal{M}}[i]$}
                \If {$SPLIT$}
                    \State  CentroidSplit($\Delta$)
                \Else
                    \State  $\Delta$.setFree
                \EndIf
            \EndFor
        \EndFor
        
        \State $\mathcal{CC}_v^{\mathcal{F}}$ =  FreeConnectedComponents($T$,$v$)
        \State sort\_ascending\_cardinality($\mathcal{CC}_v^{\mathcal{F}}$)
        \State  //Set to matter all but the CC with most tetrahedra
        \For {$i=0$ to $|\mathcal{CC}_v^{\mathcal{M}}|$-1}
            \For {$\Delta$ in $\mathcal{CC}_v^{\mathcal{M}}[i]$}
                \If {$SPLIT$}
                    \State  CentroidSplit($\Delta$)
                \Else
                    \State  $\Delta$.setMatter
                \EndIf
            \EndFor
        \EndFor
    \EndIf
\EndFor
\EndProcedure
 \end{algorithmic}
\caption{Singular Vertex Fixing. CentroidSplit is described in the text.}
 \label{Algo2}
 \end{figure}

\begin{figure*}[tbp]
    \centering
    \setlength{\tabcolsep}{1px}
    \begin{tabular}{cc}
        \def\svgwidth{0.4\textwidth}
        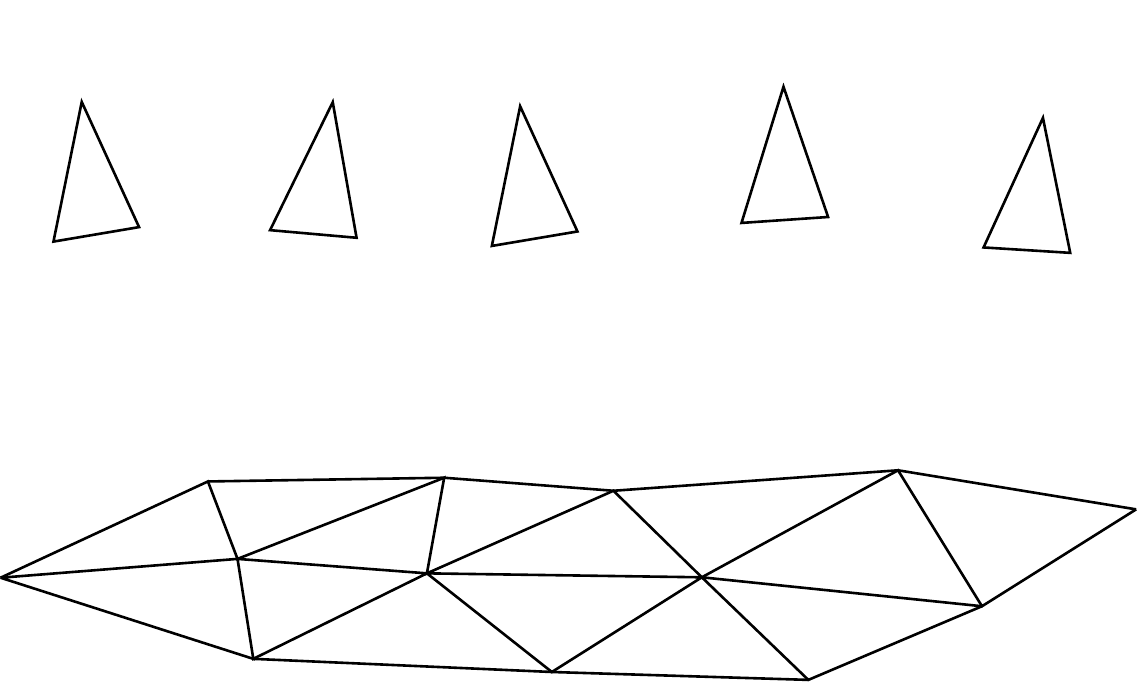
        &
        \def\svgwidth{0.4\textwidth}
        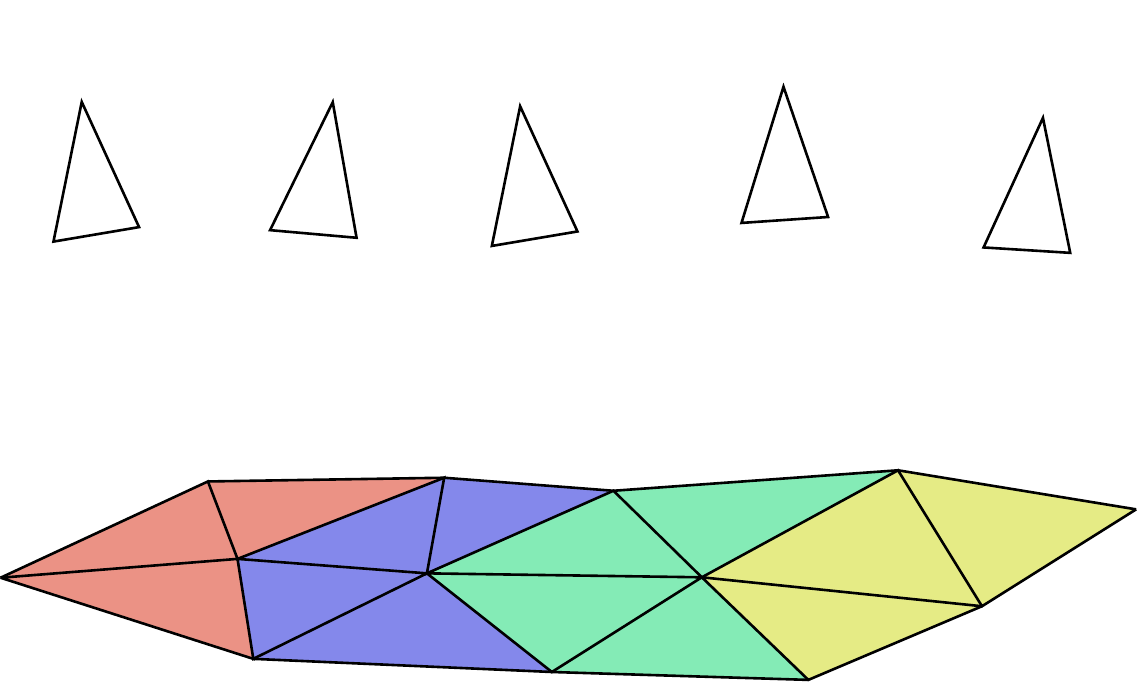\\
        (a)&(b)
    \end{tabular}
   
    \caption{The difference between standard  and facetwise camera pair selection. In the former case, \ie, (a) the choice  of cameras pairs  and the refinement consider the whole mesh model. In the facetwise selection, \ie, (b), each facet is refined by a specific pair of cameras. }
    \label{fig:facetwise}
\end{figure*}

\section{Facetwise mesh refinement}%
\label{sec:facet}
After the initial mesh is extracted by the previous volumetric method, the next step of the MVS pipeline is mesh refinement. Now, we describe the main contribution of the paper, \ie how we choose the camera pairs to compute the photo-metric error of each triangle in the Equation \eqref{eq:final}.

Given the cameras $c_1,\dots, c_N$, let's now consider the set $\mathcal{L} = \{ (c_0, c_1), (c_0, c_2),\dots, (c_{N}, c_{N-1})\}$ of all possible camera pairs that can be adopted to refine the surface mesh $\mathcal{S}$.
This set grows quadratically with the number of cameras. To make the facetwise camera selection process computationally affordable, we select a subset $\mathcal{L}^{\text{simpl}}\subset\mathcal{L}$ such that each camera is paired with two others cameras that share the highest number of reconstructed 3D points (according to the visibility from the depth map fusion). Note that selecting such pairs of cameras is computationally cheap.

The key idea of our method is to further restrict the set of pairs $\mathcal{L}^{\text{simpl}}$ to a subset that is limited but, at the same time, provides evenly distributed facet coverage, \ie, such that as many as possible facets are covered by a similar amount of camera pairs. This makes the refinement process spatially coherent, it avoids accumulating too much gradient in some regions (that minimize the term $E_{\textrm{photo}}$  and avoids leaving some region of the surfaces unseen. 

Fig. \ref{fig:facetwise} illustrates the difference between the classical approach  \ref{fig:facetwise}(a) and the proposed method  \ref{fig:facetwise}(b). The former simply use the $(i,j)$ camera pairs (colored arrows in Fig. \ref{fig:facetwise}(a)) to compute the gradient contribution of Equation \eqref{eq:final}  for the whole visible mesh. In our case, \ie, the latter, we choose the pair of cameras we use to compute the gradients for each mesh facet. For the sake of clarity, the figure shows an example of a planar surface, but our method is tailored to solve the camera selection issue especially in the presence of uneven surfaces. In these cases, if using pairs of cameras defined a priori, as in the classical methods, we cannot explicitly handle the local irregularity of the facet's orientations, and their possibly problematic visibility. This problem is addressed by our method, instead, that lets each facet choose the best two cameras for its refinement, according to its visibility information.

The problem of choosing the best set of camera pairs for each facet, can be formulated as a facet labeling. 
The algorithm looks for the most convenient labeling of the mesh facets according to visibility and spatial smoothness through a Markov Random Field. The label selected by this method eventually defines the camera pair we will use to refine the corresponding facet. In the following we treat symmetric pairs $(c_i, c_j)$ and $(c_j, c_i)$ as equivalent.

First, we collect the visibility of each vertex, defined according to the output of the depth-map fusion algorithm.
In the following, $\nu_{v}$ is the set of cameras in which a vertex $v$ is visible. 
For each triangular facet $f$ with vertices $v_0$, $v_1$ and $v_2$, we collect the list of visibility $\nu_{f} = \left\{ \nu_{v_0}, \nu_{v_1}, \nu_{v_2} \right\}$, which is the list of cameras viewing the facet vertices (maintaining the repetitions).

We define the energy of a labeling function $\mathbf{l}(f)$ that maps each facet f  to a $l_{ij} = (c_i,c_j)$ as:
\begin{equation}
    E(\mathbf{l}) = -
    \sum_{f\in \mathcal{F}} log\left( \phi(\mathbf{l}(f)) \right) \ - \sum_{(f,g)\in\mathcal{A}} log\left( \varphi(\mathbf{l}(f), \mathbf{l}(g)) \right)
\end{equation}
where $\mathcal{F}$ is the set of the facets and $\mathcal{A}$ is the set of adjacent facets. $\phi$ and $\varphi$ are the unary and pairwise potentials of the corresponding Markov Random Field. 

Given the list of visibility $\nu_{f}$ and a label $\mathbf{l}(f))$ defining a pair of cameras $(c_A, c_B)$, let us call $O_f$ as the number of occurrences of the camera $c_A$ plus the occurrences of camera $c_B$ in $\nu_f$. If one of the two cameras does not appear in  $\nu_f$ we set $O_f=0$. Then, we choose 
\begin{align}
\phi(\mathbf{l}(f)) = 
\begin{cases}
\frac{O_f}{|\nu_{f}|}, & \text{if } 
 \mathbf{l}(f) \in \mathcal{L}^{\text{simpl}}\\
0.5 \cdot \min_f\left\{\frac{O_f}{|\nu_{f}|}\right\}, & \text{if }  
\mathbf{l}(f) \notin \mathcal{L}^{\text{simpl}} 
\end{cases}
\end{align}
Indeed, if the cameras $c_A$ and $c_B$ appears frequently in the visibility list, without being filtered out in the depth fusion step, then it is  likely that the quality of the facet visibility from those camera is high. As a consequence, an effective gradient contribution can be  computed.
In very rare cases $\nu_f = \varnothing$, \ie $f$ is only visible by cameras not in $\mathcal{L}^{\text{simpl}}$, then  we fix  $\phi(\mathbf{l}(f)) = 5\cdot 10^{8}$ to avoid issues during the MRF optimization.

The pairwise $\varphi$ is instead a Potts function enforcing a smooth labeling, where $\varphi(\mathbf{l}(f), \mathbf{l}(g)) = 0.9$ $\text{if } \mathbf{l}(f) \neq \mathbf{l}(g))$ and $\varphi(\mathbf{l}(f), \mathbf{l}(g)) = 0.1$ otherwise.

The initial label $\bar{\mathbf{l}}(f)$ of each facet is given by $\bar{\mathbf{l}}(f) = argmin\{\phi(\mathbf{l}(f))\}$. In case two labels get the same score we choose the one with the camera with lowest index.

Once the energy function $E$ is minimized, we obtain a lookup table that connects the facets and the corresponding pairs of cameras. 
We then proceed with the actual refinement process. 
For each pair of cameras $(c_i,c_j)\in \mathcal{L}^{\text{simpl}}$ we compute the gradient contribution $\nabla \mathcal{E}^{im}_{ij}(x)$ in Eq. \eqref{eq:final} carried just by the facets labeled as $l_{ij}$.
Differently from the state-of-the-art, which computes the gradient for all the facets, we refine each facet using only the pair of cameras providing the best visibility.

\subsection{Implementation details}
We implemented the refinement process with C++ and OpenGL. We use GLSL to compute the gradient contributions in the shaders, and we exploit shadow mapping to efficiently compute the mesh visibility. In our implementation, we adopted the occlusion masking proposed in \cite{romanoni2019mesh}.

The current software implements the proposed camera pairs selection step but, being still under development, it still relies upon the classical mesh refinement OpenGL infrastructure therefore it does not exploit fully the significant speed-up that choosing a single pair of cameras for each facet can bring. 
Indeed, for each camera pairs, it is sufficient to activate and render the corresponding facets instead of the entire mesh. 

In the following, we provide a brief complexity analysis of the rendering process, since it is the most demanding part of the algorithm. We marginalize out the gradients collection step is common and essentially equals in the classic and the proposed solutions.
With the classical approach, given that for each of the $N$ cameras we select $K$ cameras to compare with, and given that the mesh has $F$ number of facets, we render each visible facet for each pair of cameras so $\mathcal{O}(N \cdot K \cdot F) = \mathcal{O}(N \cdot F)$.
In our case, each facet is rendered just for the camera pair selected and so the rendering stage is  $\mathcal{O}(F \cdots K ) = \mathcal{O}(F)$.

\section{Experiments}
\label{intro}
To evaluate the effectiveness of our approach we applied our pipeline in the same 12 sequences of the DTU dataset \cite{jensen2014large}, as in \cite{li2016efficient} and \cite{romanoni2019mesh}, since these are the closest and most recent methods. In particular they choose the sequences 4, 6, 15, 18, 24, 36, 63, 106, 110, 114, 118, 122. The first seven sequences consist of 49 images, the remaining by 64  images. In both cases the image resolution is 1600x1200px.
We adopted COLMAP \cite{schonberger2016structure,schonberger2016pixelwise} to compute the camera calibration, the dense map, and their fusion into a dense 3D point cloud. We adopted the procedure described in Section \ref{sec:manifold} to extract a manifold mesh which is refined according to Section \ref{sec:facet}.

In Table \ref{result_dtu} we compare the result of the proposed approach (last column) with some of the state-of-the-art and reference methods for 3D mesh reconstruction. 
We followed the methodology described in \cite{jensen2014large} to compute accuracy and completeness, expressed in mm.
Our method outperforms all the others in all the metrics except for completeness where the method of Campbell \etal \cite{campbell2008using} is better.
This table shows that the initial manifold mesh we estimated provides a good initialization, however, the refinement can significantly improve its quality. 
The facetwise refinement improves the quality of the refinement process: having the gradient contributions computed at each facet from the best pair of camera helps to evolve the surface evenly through gradient descent.

\begin{figure*}[tbp]
    \centering
    \includegraphics[width=0.85\textwidth,height=0.4\textwidth]{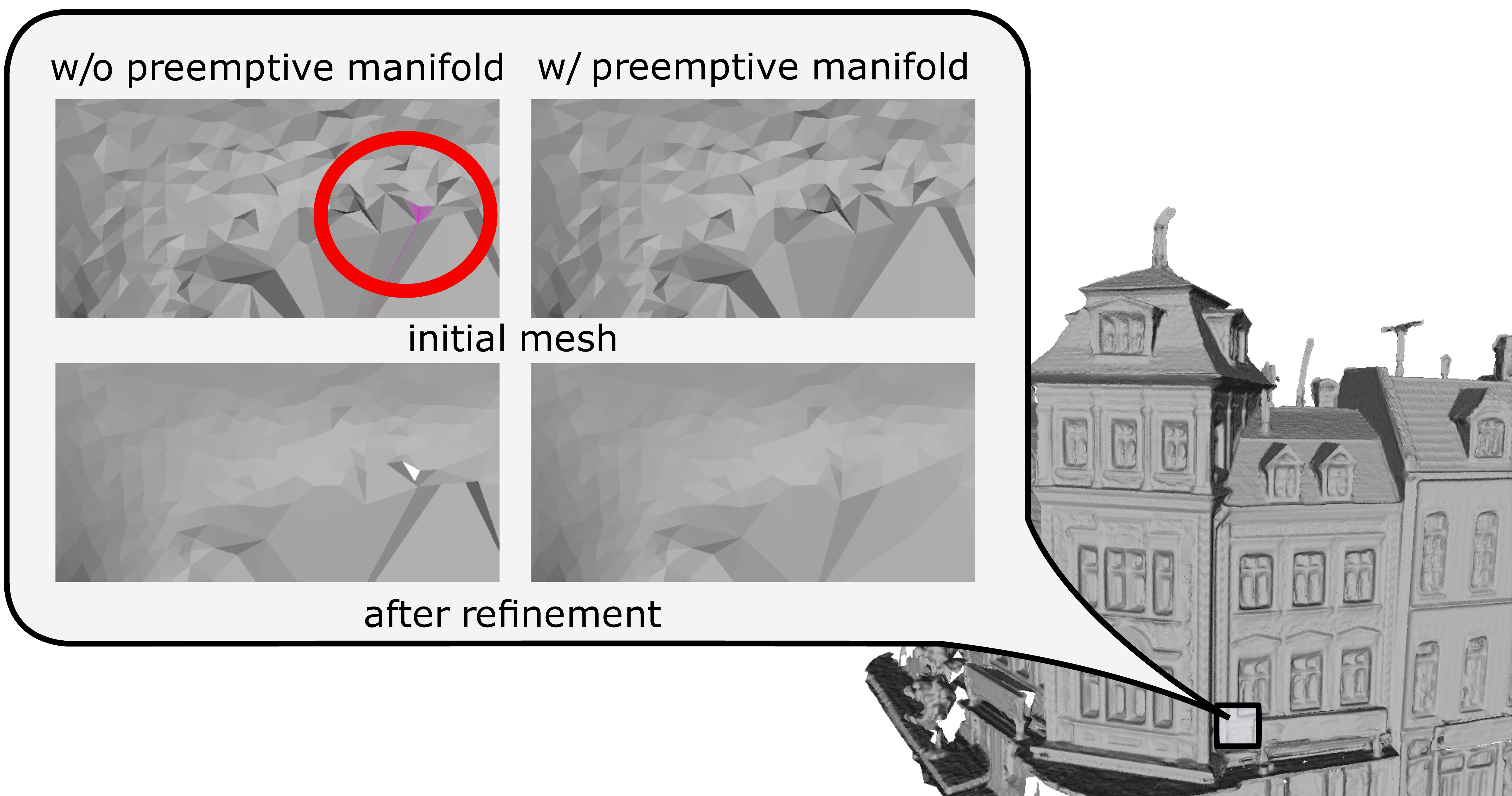}
    \caption{The effect of preemptive manifold fixing (on the right) instead of vertex split (on the left). The purple facets were not manifold. the refinement from the proposed initialization converges to a more reasonable solution. }
    \label{fig:nonman}
\end{figure*}

In Table \ref{result_dtu} we also demonstrate the importance of both steps proposed in this paper. Using just vertex splitting to fix the non-manifoldness (column w/o manif) produces inconsistencies about the mesh visibility that lead to a higher error than our method. 
Moreover, our facetwise mesh refinement achieves better performance than the classic method (w/facet column) that refines the entire mesh for each camera pairs similar to \cite{vu_et_al_2012,li2016efficient}.
In Table \ref{tab:non_man}, we further assess the effects of the preemptive singular vertex removal on the final model. Using the proposed approach leads us to reduce on average 90\% of the singular vertices appearing after the meshing step. 
Fig. \ref{fig:nonman} illustrates how refinement without the proposed approach can lead to artifacts.
Finally, Fig. \ref{fig:dtu_rec} shows the reconstructed meshes of the DTU sequences.

Table \ref{tab:epfl} illustrate the results of our method on the EPFL dataset \cite{strecha2008}. We follow the evaluation procedure in \cite{hu2012least}, \ie we show the amount of L1 depth errors below 2cm and 10 cm over the total number of valid ground-truth depths. In the comparison, we list the methods reported in \cite{schonberger2016pixelwise}. Most of them focus only on depth map estimation neglecting the fusion and meshing steps that can introduce errors. Nevertheless, our method is among the top-performing methods.

\begin{table*}[tbp]
\centering
\caption{Result on DTU Dataset. The metrics represents accuracy and completeness errors in mm (the smaller the better).}
    \setlength{\tabcolsep}{4px}
    \label{fig:pipeline}

\begin{tabular}{lccccccccc}
&\cite{campbell2008using} & \cite{furukawa2009reconstructing} & \cite{tola2012efficient} & \cite{li2016efficient} & \cite{romanoni2019mesh} & init & w/manif & w/facet&our  \\
\hline
mean acc.&1.8857 & 0.9126 & 0.4455 & 0.4245 & \textbf{0.4087} & 0.4669 & 0.4116 & 0.4298&\textbf{0.4092} \\ 
median acc.&0.7490 & 0.3973 & 0.2256 & 0.2193 & 0.2150 & 0.2195 & 0.2121 & 0.2113&\textbf{0.2067} \\ 
mean compl.&\textbf{0.4213} & 0.5330 & 0.7317 & 0.7252 & 0.7237 & 0.5200 & 0.5044 & 0.5005&0.4958 \\ 
median compl.&\textbf{0.2856} & 0.3517 & 0.3511 & 0.3426 & 0.3413 & 0.3292 & 0.3039 & 0.3010&0.2980 \\ 
Average score&0.8354 & 0.5486 & 0.4385 & 0.4279 & 0.4221 & 0.3839 & 0.3580 & 0.3607&\textbf{0.3524} \\ 
\end{tabular}
\label{result_dtu}
\end{table*}

\begin{table}[tbp]
\caption{Number of singular vertex with and without preemptive manifold fixing}
    \centering
    \setlength{\tabcolsep}{1px}
\begin{tabular}{p{50pt}cccccccccccc}
sequence & 4 & 6 & 15 & 18 & 24 & 36 & 63 & 106 & 110 & 114 & 118 & 122 \\ 
\hline
w/o  pre. fixing  & 52 & 84 & 69 & 585 & 122 & 219 & 41 & 99 & 74 & 24 & 54 & 47 \\ 
w/ pre. fixing  & 10 & 10 & 8 & 65 & 14 & 14 & 0 & 10 & 9 & 1 & 2 & 5 \\ 
\% avoided & 80.8 & 88.1 & 88.4 & 88.9 & 88.5 & 93.6 & 100 & 89.9 & 87.8 & 95.8 & 96.3 & 89.4 \\ 
\end{tabular}
\label{tab:non_man}
\end{table}

\begin{figure*}[tbp]
    \centering
    \setlength{\tabcolsep}{1px}
    \begin{tabular}{cccc}
        \includegraphics[width=0.23\textwidth]{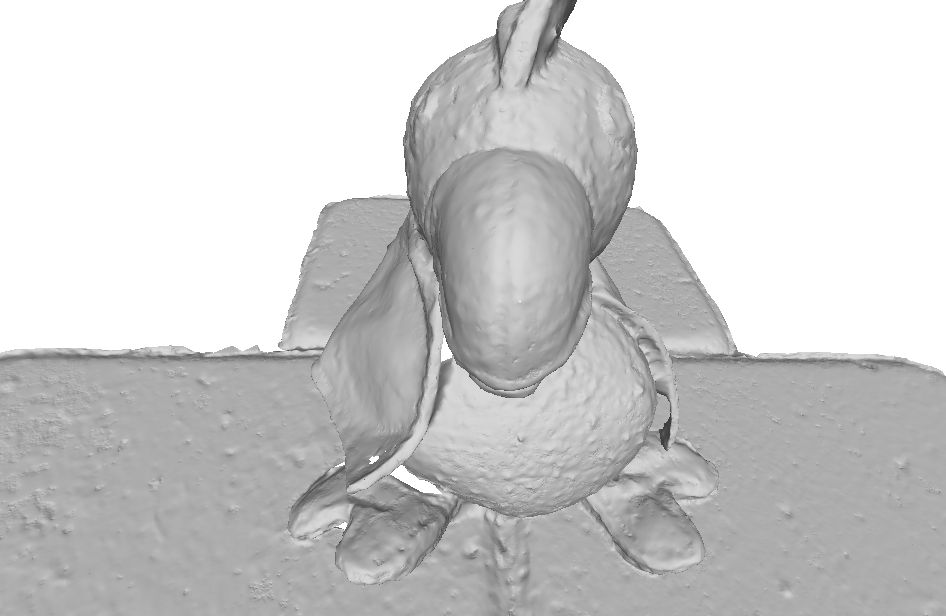}&
        \includegraphics[width=0.23\textwidth]{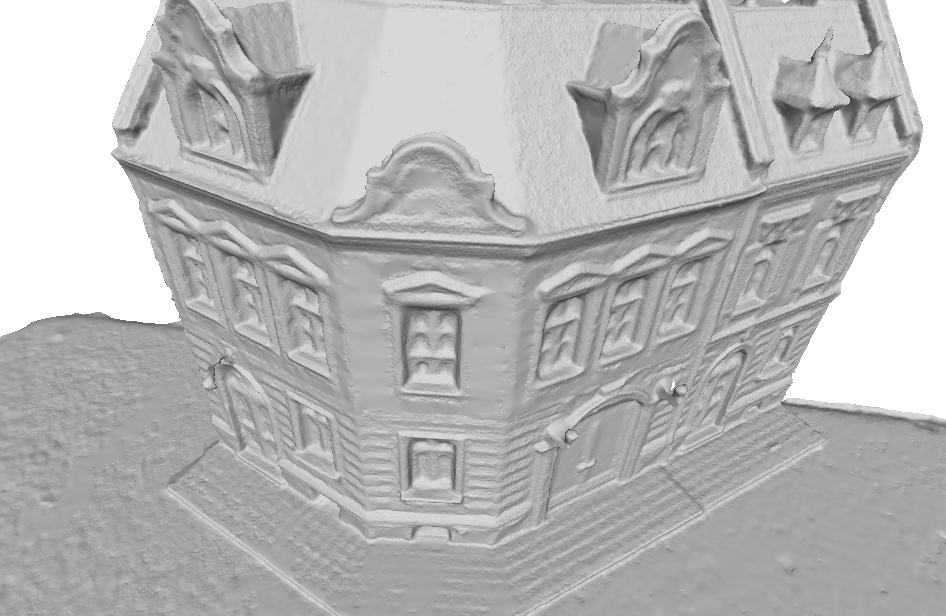}&
        \includegraphics[width=0.23\textwidth]{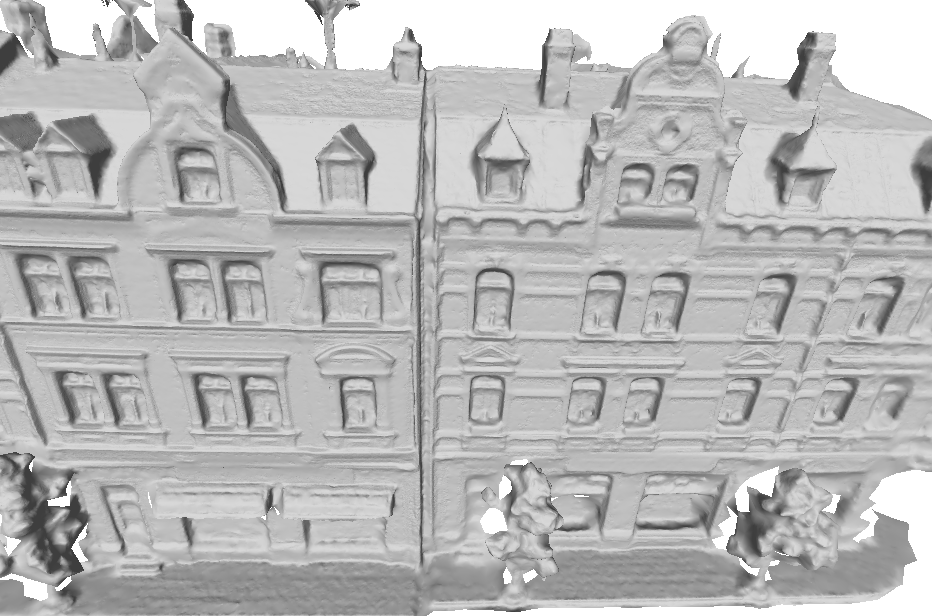}&
        \includegraphics[width=0.23\textwidth]{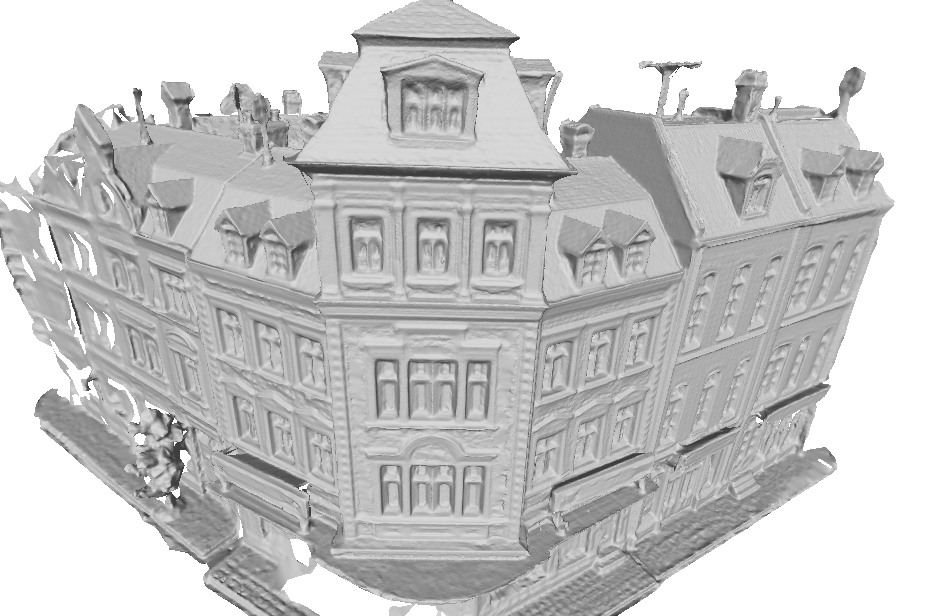}\\
        seq. 4 & seq. 6 & seq. 15 & seq. 18\\
        \includegraphics[width=0.23\textwidth]{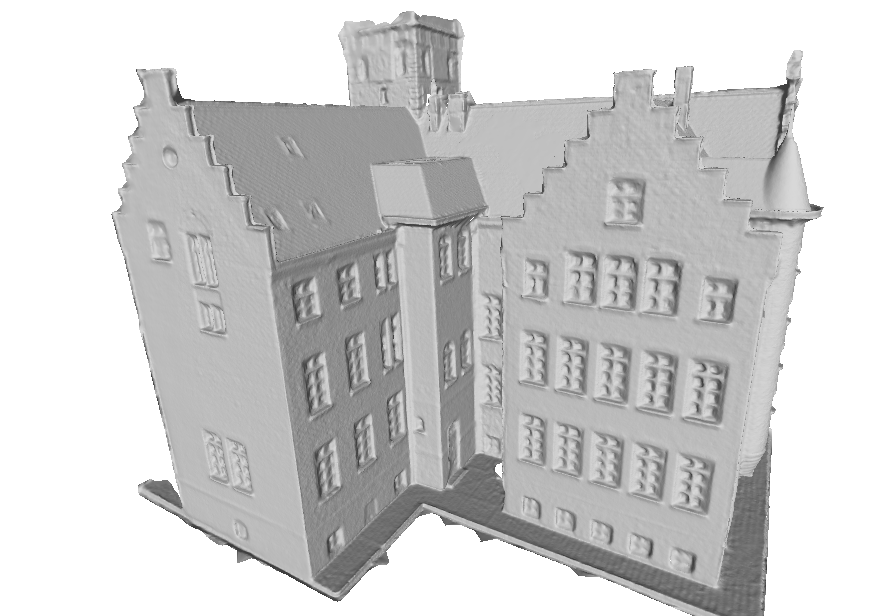}&
        \includegraphics[width=0.23\textwidth]{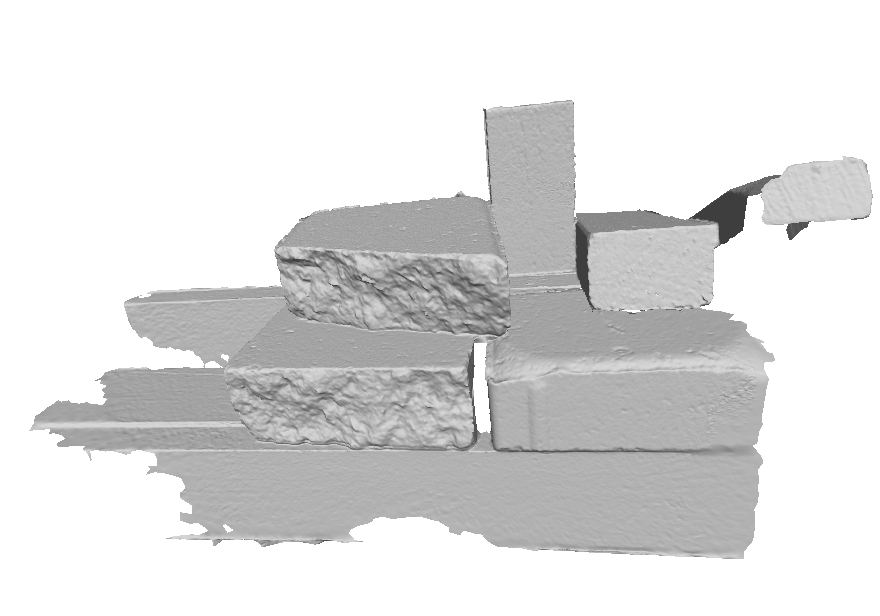}&
        \includegraphics[width=0.23\textwidth]{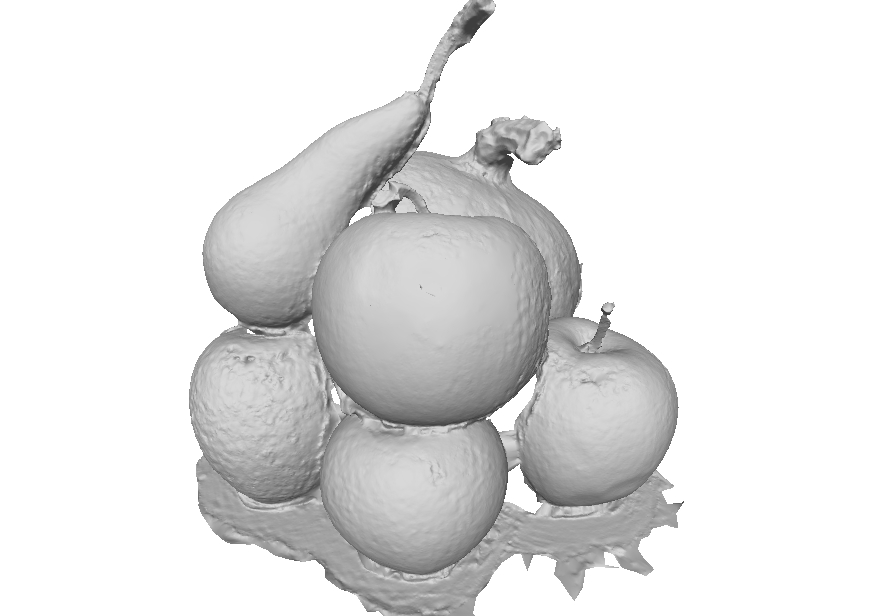}&
        \includegraphics[width=0.23\textwidth]{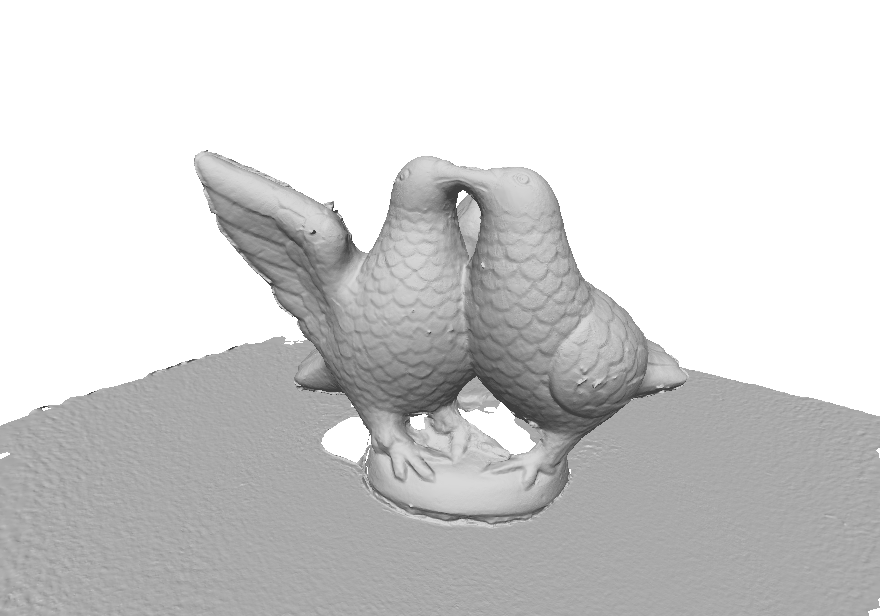}\\
        seq. 24 & seq. 36 & seq. 63 & seq. 106\\
        \includegraphics[width=0.23\textwidth]{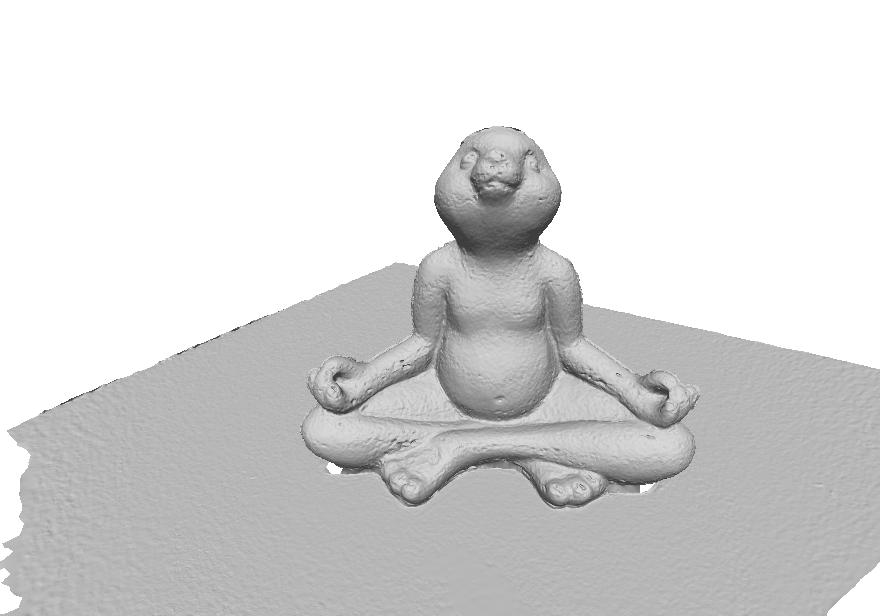}&
        \includegraphics[width=0.23\textwidth]{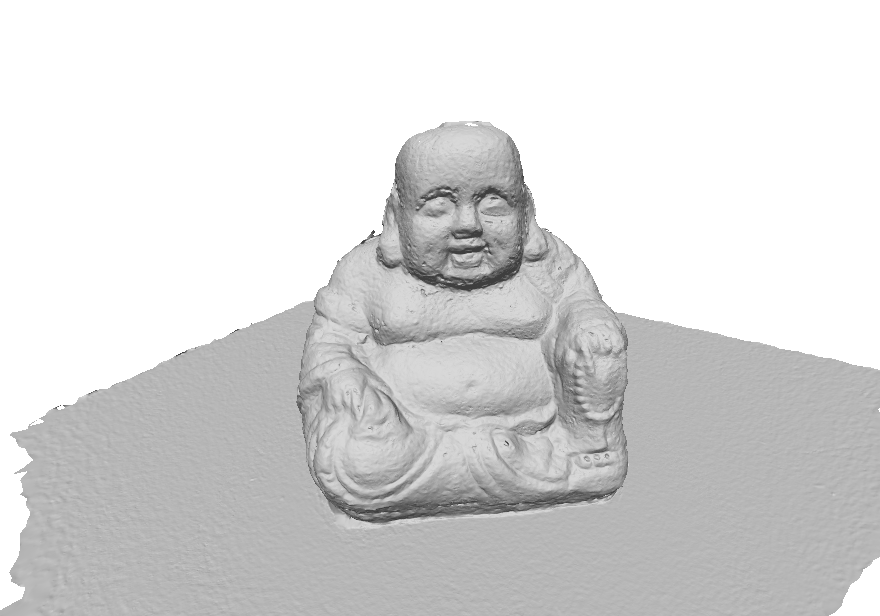}&
        \includegraphics[width=0.23\textwidth]{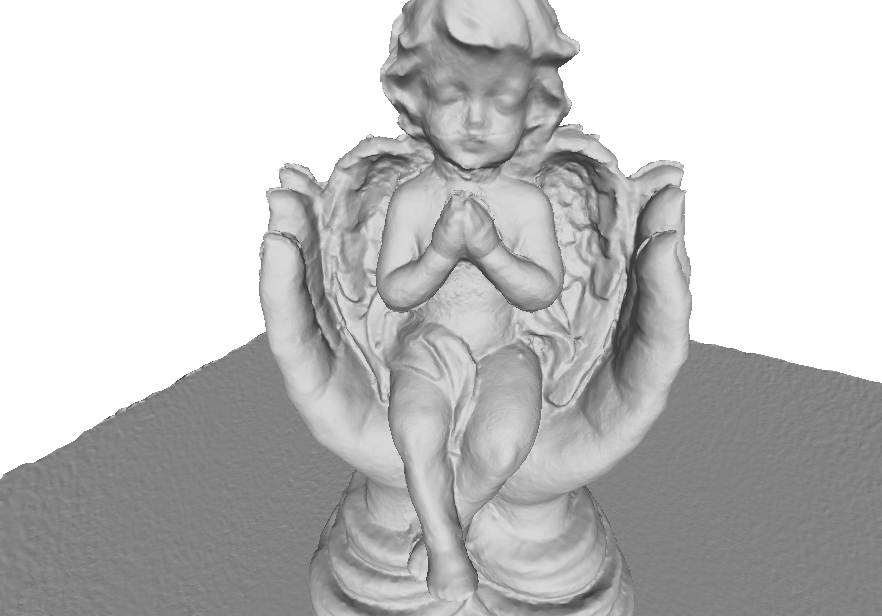}&
        \includegraphics[width=0.23\textwidth]{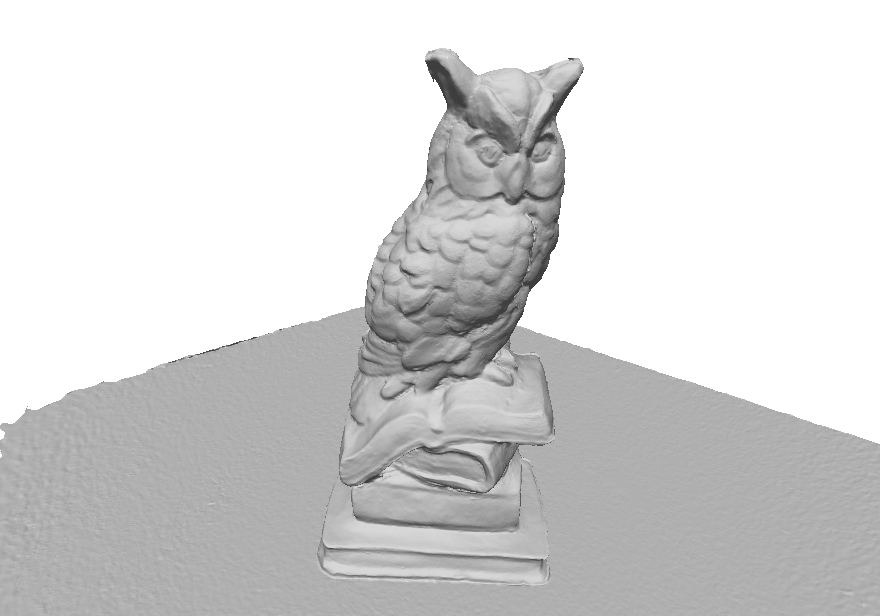}\\
        seq. 110 & seq. 114 & seq. 118 & seq. 122\\
    \end{tabular}
   
    \caption{Reconstructions of the DTU sequences. }
    \label{fig:dtu_rec}
\end{figure*}

\begin{table}[tbp]
	\caption{Results on the Epfl dataset}
    \centering
    \setlength{\tabcolsep}{3px}
	\begin{tabular}{lccccc|cccc}
	&\multicolumn{5}{c}{Depth-map output}&
	\multicolumn{4}{c}{Mesh output}\\
		&\cite{zheng14joint}&
		\cite{hu2012least}&
		\cite{fu10}&
		\cite{Galliani_2015_ICCV}&
		\cite{schonberger2016pixelwise}&
		\cite{zaharescu2007transformesh}&
		\cite{tyle_ek2010refinement}&
		\cite{jancosek2011multi}&
		our\\
		\hline
		\multicolumn{10}{l}{Fountain}\\
		2cm&0.769&0.754&0.731&0.693&\textbf{0.827}&0.712&0.732&\textbf{0.824}&\textbf{0.825}\\
		10cm&0.929&0.930&0.838&0.838&\textbf{0.975}&0.832&0.822&\textbf{0.973}&0.946\\
		\hline
		\multicolumn{10}{l}{HerzJesu}\\
		2cm &0.650 &0.649&0.646&0.283&\textbf{0.691}&0.220&0.658&\textbf{0.739}&\textbf{0.738}\\
		10cm &0.844&0.848&0.836&0.455&\textbf{0.931}&0.501&0.852&\textbf{0.923}&\textbf{0.928}\\
	\end{tabular}
	\label{tab:epfl}
\end{table}	
\section{Conclusions}
\label{sec:conclusion}
In this paper, we proposed a Multi-View Stereo pipeline to reconstruct accurate mesh models out of a set of images. 
We focus on mesh refinement, which is the last step of the pipeline and is relevant to achieve high accuracy reconstructions.
The first contribution is an extension of one of the widespread meshing steps so to preemptively avoid that singular vertices occur (the average drop is 90\%).
In this way, we obtain a manifold mesh more easily, diminishing the number of post-processing vertices splits that often cause small artifacts on the surface after refinement.
As a second contribution, we reformulate the refinement approach that usually adopts a subset of camera pairs to evolve the entire mesh. We proposed that each facet chooses which camera pair has the best visibility and therefore is better suited for its refinement. 
We demonstrated that both contributions improve the results of the state-of-the-art.

\bibliographystyle{IEEEtranS}
\bibliography{biblio.bib}

\end{document}

%% file: images/whymanifold03.pdf_tex
\begingroup%
  \makeatletter%
  \providecommand\color[2][]{%
    \errmessage{(Inkscape) Color is used for the text in Inkscape, but the package 'color.sty' is not loaded}%
    \renewcommand\color[2][]{}%
  }%
  \providecommand\transparent[1]{%
    \errmessage{(Inkscape) Transparency is used (non-zero) for the text in Inkscape, but the package 'transparent.sty' is not loaded}%
    \renewcommand\transparent[1]{}%
  }%
  \providecommand\rotatebox[2]{#2}%
  \ifx\svgwidth\undefined%
    \setlength{\unitlength}{198.53668213bp}%
    \ifx\svgscale\undefined%
      \relax%
    \else%
      \setlength{\unitlength}{\unitlength * \real{\svgscale}}%
    \fi%
  \else%
    \setlength{\unitlength}{\svgwidth}%
  \fi%
  \global\let\svgwidth\undefined%
  \global\let\svgscale\undefined%
  \makeatother%
  \begin{picture}(1,0.4748106)%
    \put(0,0){\includegraphics[width=\unitlength,page=1]{images/whymanifold03.pdf}}%
    \put(0.36519671,0.02134368){\color[rgb]{0,0,0}\makebox(0,0)[lb]{\smash{$v$}}}%
    \put(0.08206263,0.14458376){\color[rgb]{0,0,0}\makebox(0,0)[lb]{\smash{$T_1$}}}%
    \put(0.28005865,0.37088846){\color[rgb]{0,0,0}\makebox(0,0)[lb]{\smash{$T_2$}}}%
    \put(0.56264035,0.23464381){\color[rgb]{0,0,0}\makebox(0,0)[lb]{\smash{$T_3$}}}%
    \put(0.82756109,0.12362979){\color[rgb]{0,0,0}\makebox(0,0)[lb]{\smash{$T_4$}}}%
    \put(0,0){\includegraphics[width=\unitlength,page=2]{images/whymanifold03.pdf}}%
  \end{picture}%
\endgroup%

%% file: images/whymanifold04.pdf_tex
\begingroup%
  \makeatletter%
  \providecommand\color[2][]{%
    \errmessage{(Inkscape) Color is used for the text in Inkscape, but the package 'color.sty' is not loaded}%
    \renewcommand\color[2][]{}%
  }%
  \providecommand\transparent[1]{%
    \errmessage{(Inkscape) Transparency is used (non-zero) for the text in Inkscape, but the package 'transparent.sty' is not loaded}%
    \renewcommand\transparent[1]{}%
  }%
  \providecommand\rotatebox[2]{#2}%
  \ifx\svgwidth\undefined%
    \setlength{\unitlength}{198.53668213bp}%
    \ifx\svgscale\undefined%
      \relax%
    \else%
      \setlength{\unitlength}{\unitlength * \real{\svgscale}}%
    \fi%
  \else%
    \setlength{\unitlength}{\svgwidth}%
  \fi%
  \global\let\svgwidth\undefined%
  \global\let\svgscale\undefined%
  \makeatother%
  \begin{picture}(1,0.4748106)%
    \put(0,0){\includegraphics[width=\unitlength,page=1]{images/whymanifold04.pdf}}%
    \put(0.36519671,0.02134368){\color[rgb]{0,0,0}\makebox(0,0)[lb]{\smash{$v$}}}%
    \put(0.08206263,0.14458376){\color[rgb]{0,0,0}\makebox(0,0)[lb]{\smash{$T_1$}}}%
    \put(0.28005865,0.37088846){\color[rgb]{0,0,0}\makebox(0,0)[lb]{\smash{$T_2$}}}%
    \put(0.56264035,0.23464381){\color[rgb]{0,0,0}\makebox(0,0)[lb]{\smash{$T_3$}}}%
    \put(0.82756109,0.12362979){\color[rgb]{0,0,0}\makebox(0,0)[lb]{\smash{$T_4$}}}%
  \end{picture}%
\endgroup%

%% file: images/whymanifold01.pdf_tex
\begingroup%
  \makeatletter%
  \providecommand\color[2][]{%
    \errmessage{(Inkscape) Color is used for the text in Inkscape, but the package 'color.sty' is not loaded}%
    \renewcommand\color[2][]{}%
  }%
  \providecommand\transparent[1]{%
    \errmessage{(Inkscape) Transparency is used (non-zero) for the text in Inkscape, but the package 'transparent.sty' is not loaded}%
    \renewcommand\transparent[1]{}%
  }%
  \providecommand\rotatebox[2]{#2}%
  \ifx\svgwidth\undefined%
    \setlength{\unitlength}{198.53668213bp}%
    \ifx\svgscale\undefined%
      \relax%
    \else%
      \setlength{\unitlength}{\unitlength * \real{\svgscale}}%
    \fi%
  \else%
    \setlength{\unitlength}{\svgwidth}%
  \fi%
  \global\let\svgwidth\undefined%
  \global\let\svgscale\undefined%
  \makeatother%
  \begin{picture}(1,0.4748106)%
    \put(0,0){\includegraphics[width=\unitlength,page=1]{images/whymanifold01.pdf}}%
    \put(0.59932969,0.03553639){\color[rgb]{0,0,0}\makebox(0,0)[lb]{\smash{$v_2$}}}%
    \put(0,0){\includegraphics[width=\unitlength,page=2]{images/whymanifold01.pdf}}%
    \put(0.36519671,0.02134368){\color[rgb]{0,0,0}\makebox(0,0)[lb]{\smash{$v_1$}}}%
    \put(0,0){\includegraphics[width=\unitlength,page=3]{images/whymanifold01.pdf}}%
    \put(0.08206263,0.14458376){\color[rgb]{0,0,0}\makebox(0,0)[lb]{\smash{$T_1$}}}%
    \put(0.28005865,0.37088846){\color[rgb]{0,0,0}\makebox(0,0)[lb]{\smash{$T_2$}}}%
    \put(0.56264035,0.23464381){\color[rgb]{0,0,0}\makebox(0,0)[lb]{\smash{$T_3$}}}%
    \put(0.82756109,0.12362979){\color[rgb]{0,0,0}\makebox(0,0)[lb]{\smash{$T_4$}}}%
  \end{picture}%
\endgroup%

%% file: images/whymanifold02.pdf_tex
\begingroup%
  \makeatletter%
  \providecommand\color[2][]{%
    \errmessage{(Inkscape) Color is used for the text in Inkscape, but the package 'color.sty' is not loaded}%
    \renewcommand\color[2][]{}%
  }%
  \providecommand\transparent[1]{%
    \errmessage{(Inkscape) Transparency is used (non-zero) for the text in Inkscape, but the package 'transparent.sty' is not loaded}%
    \renewcommand\transparent[1]{}%
  }%
  \providecommand\rotatebox[2]{#2}%
  \ifx\svgwidth\undefined%
    \setlength{\unitlength}{198.53668213bp}%
    \ifx\svgscale\undefined%
      \relax%
    \else%
      \setlength{\unitlength}{\unitlength * \real{\svgscale}}%
    \fi%
  \else%
    \setlength{\unitlength}{\svgwidth}%
  \fi%
  \global\let\svgwidth\undefined%
  \global\let\svgscale\undefined%
  \makeatother%
  \begin{picture}(1,0.4748106)%
    \put(0,0){\includegraphics[width=\unitlength,page=1]{images/whymanifold02.pdf}}%
    \put(0.59932969,0.03553639){\color[rgb]{0,0,0}\makebox(0,0)[lb]{\smash{$v_2$}}}%
    \put(0,0){\includegraphics[width=\unitlength,page=2]{images/whymanifold02.pdf}}%
    \put(0.36519671,0.02134368){\color[rgb]{0,0,0}\makebox(0,0)[lb]{\smash{$v_1$}}}%
    \put(0.08206263,0.14458376){\color[rgb]{0,0,0}\makebox(0,0)[lb]{\smash{$T_1$}}}%
    \put(0.28005865,0.37088846){\color[rgb]{0,0,0}\makebox(0,0)[lb]{\smash{$T_2$}}}%
    \put(0.56264035,0.23464381){\color[rgb]{0,0,0}\makebox(0,0)[lb]{\smash{$T_3$}}}%
    \put(0.82756109,0.12362979){\color[rgb]{0,0,0}\makebox(0,0)[lb]{\smash{$T_4$}}}%
  \end{picture}%
\endgroup%

%% file: images/connected_non-manif_init.pdf_tex
\begingroup%
  \makeatletter%
  \providecommand\color[2][]{%
    \errmessage{(Inkscape) Color is used for the text in Inkscape, but the package 'color.sty' is not loaded}%
    \renewcommand\color[2][]{}%
  }
  \providecommand\rotatebox[2]{#2}%
  \newcommand*\fsize{\dimexpr\f@size pt\relax}%
  \newcommand*\lineheight[1]{\fontsize{\fsize}{#1\fsize}\selectfont}%
  \ifx\svgwidth\undefined%
    \setlength{\unitlength}{143.58981636bp}%
    \ifx\svgscale\undefined%
      \relax%
    \else%
      \setlength{\unitlength}{\unitlength * \real{\svgscale}}%
    \fi%
  \else%
    \setlength{\unitlength}{\svgwidth}%
  \fi%
  \global\let\svgwidth\undefined%
  \global\let\svgscale\undefined%
  \makeatother%
  \begin{picture}(1,0.8194705)%
    \lineheight{1}%
    \setlength\tabcolsep{0pt}%
    \put(0,0){\includegraphics[width=\unitlength,page=1]{images/connected_non-manif_init.pdf}}%
    \put(0.38586284,0.03366476){\color[rgb]{0,0,0}\makebox(0,0)[lt]{\lineheight{1.25}\smash{\begin{tabular}[t]{l}$v$\end{tabular}}}}%
  \end{picture}%
\endgroup%

%% file: images/connected_non-manif.pdf_tex
\begingroup%
  \makeatletter%
  \providecommand\color[2][]{%
    \errmessage{(Inkscape) Color is used for the text in Inkscape, but the package 'color.sty' is not loaded}%
    \renewcommand\color[2][]{}%
  }%
  \providecommand\transparent[1]{%
    \errmessage{(Inkscape) Transparency is used (non-zero) for the text in Inkscape, but the package 'transparent.sty' is not loaded}%
    \renewcommand\transparent[1]{}%
  }%
  \providecommand\rotatebox[2]{#2}%
  \newcommand*\fsize{\dimexpr\f@size pt\relax}%
  \newcommand*\lineheight[1]{\fontsize{\fsize}{#1\fsize}\selectfont}%
  \ifx\svgwidth\undefined%
    \setlength{\unitlength}{143.58981636bp}%
    \ifx\svgscale\undefined%
      \relax%
    \else%
      \setlength{\unitlength}{\unitlength * \real{\svgscale}}%
    \fi%
  \else%
    \setlength{\unitlength}{\svgwidth}%
  \fi%
  \global\let\svgwidth\undefined%
  \global\let\svgscale\undefined%
  \makeatother%
  \begin{picture}(1,0.8194705)%
    \lineheight{1}%
    \setlength\tabcolsep{0pt}%
    \put(0,0){\includegraphics[width=\unitlength,page=1]{images/connected_non-manif.pdf}}%
    \put(0.38586284,0.03366476){\color[rgb]{0,0,0}\makebox(0,0)[lt]{\lineheight{1.25}\smash{\begin{tabular}[t]{l}$v$\end{tabular}}}}%
  \end{picture}%
\endgroup%

%% file: images/connected_manif.pdf_tex
\begingroup%
  \makeatletter%
  \providecommand\color[2][]{%
    \errmessage{(Inkscape) Color is used for the text in Inkscape, but the package 'color.sty' is not loaded}%
    \renewcommand\color[2][]{}%
  }%
  \providecommand\transparent[1]{%
    \errmessage{(Inkscape) Transparency is used (non-zero) for the text in Inkscape, but the package 'transparent.sty' is not loaded}%
    \renewcommand\transparent[1]{}%
  }%
  \providecommand\rotatebox[2]{#2}%
  \newcommand*\fsize{\dimexpr\f@size pt\relax}%
  \newcommand*\lineheight[1]{\fontsize{\fsize}{#1\fsize}\selectfont}%
  \ifx\svgwidth\undefined%
    \setlength{\unitlength}{143.58981636bp}%
    \ifx\svgscale\undefined%
      \relax%
    \else%
      \setlength{\unitlength}{\unitlength * \real{\svgscale}}%
    \fi%
  \else%
    \setlength{\unitlength}{\svgwidth}%
  \fi%
  \global\let\svgwidth\undefined%
  \global\let\svgscale\undefined%
  \makeatother%
  \begin{picture}(1,0.8194705)%
    \lineheight{1}%
    \setlength\tabcolsep{0pt}%
    \put(0,0){\includegraphics[width=\unitlength,page=1]{images/connected_manif.pdf}}%
    \put(0.51321478,0.00617161){\color[rgb]{0,0,0}\makebox(0,0)[lt]{\lineheight{1.25}\smash{\begin{tabular}[t]{l}$v$\end{tabular}}}}%
    \put(0,0){\includegraphics[width=\unitlength,page=2]{images/connected_manif.pdf}}%
  \end{picture}%
\endgroup%

%% file: images/facetwise01.pdf_tex
\begingroup%
  \makeatletter%
  \providecommand\color[2][]{%
    \errmessage{(Inkscape) Color is used for the text in Inkscape, but the package 'color.sty' is not loaded}%
    \renewcommand\color[2][]{}%
  }%
  \providecommand\transparent[1]{%
    \errmessage{(Inkscape) Transparency is used (non-zero) for the text in Inkscape, but the package 'transparent.sty' is not loaded}%
    \renewcommand\transparent[1]{}%
  }%
  \providecommand\rotatebox[2]{#2}%
  \newcommand*\fsize{\dimexpr\f@size pt\relax}%
  \newcommand*\lineheight[1]{\fontsize{\fsize}{#1\fsize}\selectfont}%
  \ifx\svgwidth\undefined%
    \setlength{\unitlength}{327.40719773bp}%
    \ifx\svgscale\undefined%
      \relax%
    \else%
      \setlength{\unitlength}{\unitlength * \real{\svgscale}}%
    \fi%
  \else%
    \setlength{\unitlength}{\svgwidth}%
  \fi%
  \global\let\svgwidth\undefined%
  \global\let\svgscale\undefined%
  \makeatother%
  \begin{picture}(1,0.59941129)%
    \lineheight{1}%
    \setlength\tabcolsep{0pt}%
    \put(0,0){\includegraphics[width=\unitlength,page=1]{images/facetwise01.pdf}}%
    \put(0.06430533,0.53240881){\color[rgb]{0,0,0}\makebox(0,0)[lt]{\lineheight{1.25}\smash{\begin{tabular}[t]{l}$C_1$\end{tabular}}}}%
    \put(0.28728361,0.53910314){\color[rgb]{0,0,0}\makebox(0,0)[lt]{\lineheight{1.25}\smash{\begin{tabular}[t]{l}$C_2$\end{tabular}}}}%
    \put(0.45389027,0.54488811){\color[rgb]{0,0,0}\makebox(0,0)[lt]{\lineheight{1.25}\smash{\begin{tabular}[t]{l}$C_3$\end{tabular}}}}%
    \put(0.66677671,0.54835913){\color[rgb]{0,0,0}\makebox(0,0)[lt]{\lineheight{1.25}\smash{\begin{tabular}[t]{l}$C_4$\end{tabular}}}}%
    \put(0.88350612,0.53216117){\color[rgb]{0,0,0}\makebox(0,0)[lt]{\lineheight{1.25}\smash{\begin{tabular}[t]{l}$C_5$\end{tabular}}}}%
    \put(0,0){\includegraphics[width=\unitlength,page=2]{images/facetwise01.pdf}}%
  \end{picture}%
\endgroup%

%% file: images/facetwise_prop.pdf_tex
\begingroup%
  \makeatletter%
  \providecommand\color[2][]{%
    \errmessage{(Inkscape) Color is used for the text in Inkscape, but the package 'color.sty' is not loaded}%
    \renewcommand\color[2][]{}%
  }%
  \providecommand\transparent[1]{%
    \errmessage{(Inkscape) Transparency is used (non-zero) for the text in Inkscape, but the package 'transparent.sty' is not loaded}%
    \renewcommand\transparent[1]{}%
  }%
  \providecommand\rotatebox[2]{#2}%
  \newcommand*\fsize{\dimexpr\f@size pt\relax}%
  \newcommand*\lineheight[1]{\fontsize{\fsize}{#1\fsize}\selectfont}%
  \ifx\svgwidth\undefined%
    \setlength{\unitlength}{327.40719773bp}%
    \ifx\svgscale\undefined%
      \relax%
    \else%
      \setlength{\unitlength}{\unitlength * \real{\svgscale}}%
    \fi%
  \else%
    \setlength{\unitlength}{\svgwidth}%
  \fi%
  \global\let\svgwidth\undefined%
  \global\let\svgscale\undefined%
  \makeatother%
  \begin{picture}(1,0.59941129)%
    \lineheight{1}%
    \setlength\tabcolsep{0pt}%
    \put(0,0){\includegraphics[width=\unitlength,page=1]{images/facetwise_prop.pdf}}%
    \put(0.06430533,0.53240881){\color[rgb]{0,0,0}\makebox(0,0)[lt]{\lineheight{1.25}\smash{\begin{tabular}[t]{l}$C_1$\end{tabular}}}}%
    \put(0.28728361,0.53910314){\color[rgb]{0,0,0}\makebox(0,0)[lt]{\lineheight{1.25}\smash{\begin{tabular}[t]{l}$C_2$\end{tabular}}}}%
    \put(0.45389027,0.54488811){\color[rgb]{0,0,0}\makebox(0,0)[lt]{\lineheight{1.25}\smash{\begin{tabular}[t]{l}$C_3$\end{tabular}}}}%
    \put(0.66677671,0.54835913){\color[rgb]{0,0,0}\makebox(0,0)[lt]{\lineheight{1.25}\smash{\begin{tabular}[t]{l}$C_4$\end{tabular}}}}%
    \put(0.88350612,0.53216117){\color[rgb]{0,0,0}\makebox(0,0)[lt]{\lineheight{1.25}\smash{\begin{tabular}[t]{l}$C_5$\end{tabular}}}}%
    \put(0.56785152,0.0231961){\color[rgb]{0,0,0}\makebox(0,0)[lt]{\lineheight{1.25}\smash{\begin{tabular}[t]{l}\tiny$(C_3, C_4)$\end{tabular}}}}%
    \put(0.08534513,0.07708525){\color[rgb]{0,0,0}\makebox(0,0)[lt]{\lineheight{1.25}\smash{\begin{tabular}[t]{l}\tiny$(C_1, C_2)$\end{tabular}}}}%
    \put(0.08734227,0.11521796){\color[rgb]{0,0,0}\makebox(0,0)[lt]{\lineheight{1.25}\smash{\begin{tabular}[t]{l}\tiny$(C_1, C_2)$\end{tabular}}}}%
    \put(0.20408411,0.14767173){\color[rgb]{0,0,0}\makebox(0,0)[lt]{\lineheight{1.25}\smash{\begin{tabular}[t]{l}\tiny$(C_1, C_2)$\end{tabular}}}}%
    \put(0.2907181,0.11609424){\color[rgb]{0,0,0}\makebox(0,0)[lt]{\lineheight{1.25}\smash{\begin{tabular}[t]{l}\tiny$(C_2, C_3)$\end{tabular}}}}%
    \put(0.2355214,0.0724674){\color[rgb]{0,0,0}\makebox(0,0)[lt]{\lineheight{1.25}\smash{\begin{tabular}[t]{l}\tiny$(C_2, C_3)$\end{tabular}}}}%
    \put(0.32556817,0.03544361){\color[rgb]{0,0,0}\makebox(0,0)[lt]{\lineheight{1.25}\smash{\begin{tabular}[t]{l}\tiny$(C_2, C_3)$\end{tabular}}}}%
    \put(0.47792807,0.11378022){\color[rgb]{0,0,0}\makebox(0,0)[lt]{\lineheight{1.25}\smash{\begin{tabular}[t]{l}\tiny$(C_3, C_4)$\end{tabular}}}}%
    \put(0.59986728,0.14851405){\color[rgb]{0,0,0}\makebox(0,0)[lt]{\lineheight{1.25}\smash{\begin{tabular}[t]{l}\tiny$(C_3, C_4)$\end{tabular}}}}%
    \put(0.71440934,0.10554088){\color[rgb]{0,0,0}\makebox(0,0)[lt]{\lineheight{1.25}\smash{\begin{tabular}[t]{l}\tiny$(C_4, C_5)$\end{tabular}}}}%
    \put(0.69349129,0.04354265){\color[rgb]{0,0,0}\makebox(0,0)[lt]{\lineheight{1.25}\smash{\begin{tabular}[t]{l}\tiny$(C_4, C_5)$\end{tabular}}}}%
    \put(0.85891697,0.13133361){\color[rgb]{0,0,0}\makebox(0,0)[lt]{\lineheight{1.25}\smash{\begin{tabular}[t]{l}\tiny$(C_4, C_5)$\end{tabular}}}}%
    \put(0.4088475,0.14685371){\color[rgb]{0,0,0}\makebox(0,0)[lt]{\lineheight{1.25}\smash{\begin{tabular}[t]{l}\tiny$(C_2, C_3)$\end{tabular}}}}%
    \put(0.45817699,0.06137676){\color[rgb]{0,0,0}\makebox(0,0)[lt]{\lineheight{1.25}\smash{\begin{tabular}[t]{l}\tiny$(C_3, C_4)$\end{tabular}}}}%
  \end{picture}%
\endgroup%

%% file: root.bbl
\begin{thebibliography}{10}
\providecommand{\url}[1]{#1}
\csname url@samestyle\endcsname
\providecommand{\newblock}{\relax}
\providecommand{\bibinfo}[2]{#2}
\providecommand{\BIBentrySTDinterwordspacing}{\spaceskip=0pt\relax}
\providecommand{\BIBentryALTinterwordstretchfactor}{4}
\providecommand{\BIBentryALTinterwordspacing}{\spaceskip=\fontdimen2\font plus
\BIBentryALTinterwordstretchfactor\fontdimen3\font minus
  \fontdimen4\font\relax}
\providecommand{\BIBforeignlanguage}[2]{{%
\expandafter\ifx\csname l@#1\endcsname\relax
\typeout{** WARNING: IEEEtranS.bst: No hyphenation pattern has been}%
\typeout{** loaded for the language `#1'. Using the pattern for}%
\typeout{** the default language instead.}%
\else
\language=\csname l@#1\endcsname
\fi
#2}}
\providecommand{\BIBdecl}{\relax}
\BIBdecl

\bibitem{blaha2017semantically}
M.~Blaha, M.~Rothermel, M.~R. Oswald, T.~Sattler, A.~Richard, J.~D. Wegner,
  M.~Pollefeys, and K.~Schindler, ``Semantically informed multiview surface
  refinement,'' \emph{International Journal of Computer Vision}, 2017.

\bibitem{campbell2008using}
N.~Campbell, G.~Vogiatzis, C.~Hern{\'a}ndez, and R.~Cipolla, ``Using multiple
  hypotheses to improve depth-maps for multi-view stereo,'' \emph{Computer
  Vision--ECCV 2008}, pp. 766--779, 2008.

\bibitem{furukawa2009reconstructing}
Y.~Furukawa, B.~Curless, S.~M. Seitz, and R.~Szeliski, ``Reconstructing
  building interiors from images,'' in \emph{Computer Vision, 2009 IEEE 12th
  International Conference on}.\hskip 1em plus 0.5em minus 0.4em\relax IEEE,
  2009, pp. 80--87.

\bibitem{fu10}
Y.~Furukawa and J.~Ponce, ``Accurate, dense, and robust multiview stereopsis,''
  \emph{Pattern Analysis and Machine Intelligence, IEEE Transactions on},
  vol.~32, no.~8, pp. 1362--1376, 2010.

\bibitem{Galliani_2015_ICCV}
S.~Galliani, K.~Lasinger, and K.~Schindler, ``Massively parallel multiview
  stereopsis by surface normal diffusion,'' \emph{The IEEE International
  Conference on Computer Vision (ICCV)}, June 2015.

\bibitem{hane_et_al_09}
C.~H\"{a}ne, C.~Zach, A.~Cohen, R.~Angst, and M.~Pollefeys, ``Joint 3d scene
  reconstruction and class segmentation,'' in \emph{Computer Vision and Pattern
  Recognition (CVPR), 2013 IEEE Conference on}.\hskip 1em plus 0.5em minus
  0.4em\relax IEEE, 2013, pp. 97--104.

\bibitem{hu2012least}
X.~Hu and P.~Mordohai, ``Least commitment, viewpoint-based, multi-view
  stereo,'' in \emph{2012 Second International Conference on 3D Imaging,
  Modeling, Processing, Visualization \& Transmission}.\hskip 1em plus 0.5em
  minus 0.4em\relax IEEE, 2012, pp. 531--538.

\bibitem{jancosek2011multi}
M.~Jancosek and T.~Pajdla, ``Multi-view reconstruction preserving
  weakly-supported surfaces,'' in \emph{Computer Vision and Pattern Recognition
  (CVPR), 2011 IEEE Conference on}.\hskip 1em plus 0.5em minus 0.4em\relax
  IEEE, 2011, pp. 3121--3128.

\bibitem{jensen2014large}
R.~Jensen, A.~Dahl, G.~Vogiatzis, E.~Tola, and H.~Aan{\ae}s, ``Large scale
  multi-view stereopsis evaluation,'' in \emph{2014 IEEE Conference on Computer
  Vision and Pattern Recognition}.\hskip 1em plus 0.5em minus 0.4em\relax IEEE,
  2014, pp. 406--413.

\bibitem{jin2002variational}
H.~Jin, A.~J. Yezzi, and S.~Soatto, ``Variational multiframe stereo in the
  presence of specular reflections,'' in \emph{null}.\hskip 1em plus 0.5em
  minus 0.4em\relax IEEE, 2002, p. 626.

\bibitem{kazhdan2006poisson}
M.~Kazhdan, M.~Bolitho, and H.~Hoppe, ``Poisson surface reconstruction,'' in
  \emph{Proceedings of the fourth Eurographics symposium on Geometry
  processing}, vol.~7, 2006.

\bibitem{labatut2007efficient}
P.~Labatut, J.-P. Pons, and R.~Keriven, ``Efficient multi-view reconstruction
  of large-scale scenes using interest points, delaunay triangulation and graph
  cuts,'' in \emph{Computer Vision, 2007. ICCV 2007. IEEE 11th International
  Conference on}.\hskip 1em plus 0.5em minus 0.4em\relax IEEE, 2007, pp. 1--8.

\bibitem{lhuillier2018surface}
M.~Lhuillier, ``Surface reconstruction from a sparse point cloud by enforcing
  visibility consistency and topology constraints,'' \emph{Computer Vision and
  Image Understanding}, vol. 175, pp. 52--71, 2018.

\bibitem{lhuillier_yu2013}
M.~Lhuillier and S.~Yu, ``Manifold surface reconstruction of an environment
  from sparse structure-from-motion data,'' \emph{Computer Vision and Image
  Understanding}, vol. 117, no.~11, pp. 1628--1644, 2013.

\bibitem{li2016efficient}
S.~Li, S.~Y. Siu, T.~Fang, and L.~Quan, ``Efficient multi-view surface
  refinement with adaptive resolution control,'' in \emph{European Conference
  on Computer Vision}.\hskip 1em plus 0.5em minus 0.4em\relax Springer, 2016,
  pp. 349--364.

\bibitem{litvinov_lhiuller14}
V.~Litvinov and M.~Lhuillier, ``Incremental solid modeling from sparse
  structure-from-motion data with improved visual artifacts removal,'' in
  \emph{International Conference on Pattern Recognition (ICPR)}, 2014.

\bibitem{lorensen1987marching}
W.~E. Lorensen and H.~E. Cline, ``Marching cubes: A high resolution 3d surface
  reconstruction algorithm,'' in \emph{ACM siggraph computer graphics},
  vol.~21.\hskip 1em plus 0.5em minus 0.4em\relax ACM, 1987, pp. 163--169.

\bibitem{lovi_et_al_11}
D.~I. Lovi, N.~Birkbeck, D.~Cobzas, and M.~Jagersand, ``Incremental free-space
  carving for real-time 3d reconstruction,'' in \emph{Fifth International
  Symposium on 3D Data Processing Visualization and Transmission(3DPVT)}, 2010.

\bibitem{moulon2012adaptive}
P.~Moulon, P.~Monasse, and R.~Marlet, ``Adaptive structure from motion with a
  contrario model estimation,'' in \emph{Asian Conference on Computer
  Vision}.\hskip 1em plus 0.5em minus 0.4em\relax Springer, 2012, pp. 257--270.

\bibitem{pons2007multi}
J.-P. Pons, R.~Keriven, and O.~Faugeras, ``Multi-view stereo reconstruction and
  scene flow estimation with a global image-based matching score,''
  \emph{International Journal of Computer Vision}, vol.~72, no.~2, pp.
  179--193, 2007.

\bibitem{romanoni2017multi}
A.~Romanoni, M.~Ciccone, F.~Visin, and M.~Matteucci, ``Multi-view stereo with
  single-view semantic mesh refinement,'' in \emph{Proceedings of the IEEE
  International Conference on Computer Vision Workshops}, 2017, pp. 706--715.

\bibitem{romanoni16}
A.~Romanoni, A.~Delaunoy, M.~Pollefeys, and M.~Matteucci, ``Automatic 3d
  reconstruction of manifold meshes via delaunay triangulation and mesh
  sweeping,'' in \emph{Winter Conference on Applications of Computer Vision
  (WACV)}.\hskip 1em plus 0.5em minus 0.4em\relax IEEE, 2016.

\bibitem{romanoni15b}
A.~Romanoni and M.~Matteucci, ``Incremental reconstruction of urban
  environments by edge-points delaunay triangulation,'' in \emph{Intelligent
  Robots and Systems (IROS), 2015 IEEE/RSJ International Conference on}.\hskip
  1em plus 0.5em minus 0.4em\relax IEEE, 2015, pp. 4473--4479.

\bibitem{romanoni2019mesh}
------, ``Mesh-based camera pairs selection and occlusion-aware masking for
  mesh refinement,'' \emph{Pattern Recognition Letters}, vol. 125, pp.
  364--372, 2019.

\bibitem{savinov2015discrete}
N.~Savinov, L.~Ladicky, C.~H\"{a}ne, and M.~Pollefeys, ``Discrete optimization
  of ray potentials for semantic 3d reconstruction,'' in \emph{Computer Vision
  and Pattern Recognition (CVPR), 2015 IEEE Conference on}.\hskip 1em plus
  0.5em minus 0.4em\relax IEEE, 2015, pp. 5511--5518.

\bibitem{schonberger2016structure}
J.~L. Schonberger and J.-M. Frahm, ``Structure-from-motion revisited,'' in
  \emph{Proceedings of the IEEE Conference on Computer Vision and Pattern
  Recognition}, 2016, pp. 4104--4113.

\bibitem{schonberger2016pixelwise}
J.~L. Sch{\"o}nberger, E.~Zheng, J.-M. Frahm, and M.~Pollefeys, ``Pixelwise
  view selection for unstructured multi-view stereo,'' in \emph{European
  Conference on Computer Vision}.\hskip 1em plus 0.5em minus 0.4em\relax
  Springer, 2016, pp. 501--518.

\bibitem{shen2013accurate}
S.~Shen, ``Accurate multiple view 3d reconstruction using patch-based stereo
  for large-scale scenes.'' \emph{IEEE transactions on image processing},
  vol.~22, no.~5, pp. 1901--1914, 2013.

\bibitem{strecha2008}
C.~Strecha, W.~von Hansen, L.~Van~Gool, P.~Fua, and U.~Thoennessen, ``On
  benchmarking camera calibration and multi-view stereo for high resolution
  imagery,'' in \emph{Computer Vision and Pattern Recognition, 2008. CVPR 2008.
  IEEE Conference on}.\hskip 1em plus 0.5em minus 0.4em\relax IEEE, 2008, pp.
  1--8.

\bibitem{tola2012efficient}
E.~Tola, C.~Strecha, and P.~Fua, ``Efficient large-scale multi-view stereo for
  ultra high-resolution image sets,'' \emph{Machine Vision and Applications},
  vol.~23, no.~5, pp. 903--920, 2012.

\bibitem{tyle_ek2010refinement}
R.~Tyle\v{c}ek and R.~{\v{S}}{\'a}ra, ``Refinement of surface mesh for accurate
  multi-view reconstruction,'' \emph{International Journal of Virtual Reality},
  vol.~9, no.~1, pp. 45--54, 2010.

\bibitem{VuPhD011}
H.~H. Vu, ``Stéreo multi-vues a grànde échelleet de haute qualité,'' Ph.D.
  dissertation, Ecole des ponts Paristech, Dec 2011.

\bibitem{vu_et_al_2012}
H.~H. Vu, P.~Labatut, J.-P. Pons, and R.~Keriven, ``High accuracy and
  visibility-consistent dense multiview stereo,'' \emph{Pattern Analysis and
  Machine Intelligence, IEEE Transactions on}, vol.~34, no.~5, pp. 889--901,
  2012.

\bibitem{wardetzky2007discrete}
M.~Wardetzky, S.~Mathur, F.~K{\"a}lberer, and E.~Grinspun, ``Discrete laplace
  operators: no free lunch,'' in \emph{Symposium on Geometry processing}, 2007,
  pp. 33--37.

\bibitem{wu2011visualsfm}
C.~Wu, ``Visualsfm: A visual structure from motion system,'' 2011.

\bibitem{yao2018mvsnet}
Y.~Yao, Z.~Luo, S.~Li, T.~Fang, and L.~Quan, ``Mvsnet: Depth inference for
  unstructured multi-view stereo,'' \emph{European Conference on Computer
  Vision (ECCV)}, 2018.

\bibitem{yoon2010joint}
K.-J. Yoon, E.~Prados, and P.~Sturm, ``Joint estimation of shape and
  reflectance using multiple images with known illumination conditions,''
  \emph{International Journal of Computer Vision}, vol.~86, no. 2-3, pp.
  192--210, 2010.

\bibitem{zach2007globally}
C.~Zach, T.~Pock, and H.~Bischof, ``A globally optimal algorithm for robust
  tv-l 1 range image integration,'' in \emph{Computer Vision, 2007. ICCV 2007.
  IEEE 11th International Conference on}.\hskip 1em plus 0.5em minus
  0.4em\relax IEEE, 2007, pp. 1--8.

\bibitem{zaharescu2007transformesh}
A.~Zaharescu, E.~Boyer, and R.~Horaud, ``Transformesh: a topology-adaptive
  mesh-based approach to surface evolution,'' in \emph{Computer Vision--ACCV
  2007}.\hskip 1em plus 0.5em minus 0.4em\relax Springer, 2007, pp. 166--175.

\bibitem{zheng14joint}
E.~Zheng, E.~Dunn, V.~Jojic, and J.-M. Frahm, ``Patchmatch based joint view
  selection and depthmap estimation,'' in \emph{IEEE Conference on Computer
  Vision and Pattern Recognition}, June 2014, pp. 1510--1517.

\end{thebibliography}
